\documentclass[11pt]{article}

\usepackage{fullpage}
\usepackage{amsmath,amsfonts,amsthm,amssymb,xspace,bm, verbatim,dsfont,mathtools}
\usepackage{natbib}
\usepackage{graphicx}
\usepackage{url, algorithm, algorithmic} %url
\usepackage{color}
\usepackage{epstopdf}
\usepackage{appendix}
\usepackage{epstopdf}
\theoremstyle{plain}
\newtheorem{theorem}{Theorem}[section]
\newtheorem{lemma}[theorem]{Lemma}
\newtheorem*{lemma*}{Lemma}
\newtheorem{claim}[theorem]{Claim}
\theoremstyle{definition}
\newcommand{\rnn}{\mathbb{R}^{n \times n}}

\newcommand{\eps}{\epsilon}

\newcommand{\Var}{\operatorname{Var}}

\newcommand{\note}[1]{\marginpar{\tiny *note in TeX*}}
\newcommand{\ignore}[1]{}

\renewcommand{\phi}{\varphi}

\newcommand{\R}{\mathbb{R}}

\newcommand{\eqdef}{\stackrel{\textrm{def}}{=}}

\DeclareMathOperator*{\argmin}{argmin}

\newcommand{\expec}[1]{\mathbb{E}\left[#1\right]}
\newcommand{\prob}[1]{\mathbb{P}\left[#1\right]}

\newcommand{\So}{\Sigma^*}
\newcommand{\Uo}{U^*}

\newcommand{\ip}[2]{\langle #1, #2 \rangle}

\def\Po{P_{\Omega}}
\def\Ro{R_{\Omega}}
\def\lv{\left\vert}
\def\rv{\right\vert}
\def\lV{\left\Vert}
\def\rV{\right\Vert}

\def\T{\mathcal{T}}
\def\E{\mathcal{E}}
\def\H{\mathcal{H}}
\def\W{\mathcal{W}}

\def \wT{\widehat{\mathcal{T}}}
\def \f1q{\frac{1}{q}}

\def\pijk{p_{ijk}}

\def\dijk{\delta_{ijk}}
\def\wijk{\W_{ijk}}

\newcommand{\utp}{U^{t+1}}

\newcommand{\wutp}{\widehat{U}^{t+1}}
\def\so{\sigma^*}
\def\ftt{\frac{3}{2}}

\begin{document}
%\twocolumn[
%\icmltitle{A New Sampling Technique for Tensors}
%\icmlkeywords{tensor factorization, sparsification, tensor completion, sampling}

%\vskip 0.3in
%]
\title{A New Sampling Technique for Tensors}
\author{
Srinadh Bhojanapalli\\
{The University of Texas at Austin}\\
{bsrinadh@utexas.edu}
\and
Sujay Sanghavi\\
{The University of Texas at Austin}\\
{sanghavi@mail.utexas.edu}
}
\maketitle

\begin{abstract}
In this paper we propose new techniques to sample arbitrary third-order tensors, with an objective of speeding up tensor algorithms that have recently gained popularity in machine learning. Our main contribution is a new way to select, in a biased random way, only $O(n^{1.5}/\epsilon^2)$ of the possible $n^3$ elements while still achieving each of the three goals: \\
{\em (a) tensor sparsification}: for a tensor that has to be formed from arbitrary samples, compute very few elements to get a good spectral approximation, and for arbitrary orthogonal tensors {\em (b) tensor completion:} recover an exactly low-rank tensor from a small number of samples via alternating least squares, or  {\em (c) tensor factorization:} approximating factors of a low-rank tensor corrupted by noise. \\
Our sampling can be used along with existing tensor-based algorithms to speed them up, removing the computational bottleneck in these methods.
\end{abstract}
%\newpage

\section{Introduction}\label{sec:intro}

Tensors capture higher order relations in the data. Computing factors of higher order tensors has been of interest for a long time in chemometrics~\citep{smilde2005multi}, psychometrics, neuroscience~\citep{acar2007multiway, kolda2009tensor} and recently of increasing interest in machine learning~\citep{signoretto2014learning, yilmaz2011generalised, liu2013tensor} with applications in learning latent variable models like hidden Markov models (HMMs), Gaussian mixture models and latent Dirichlet allocation (LDA)~\citep{anandkumar2014tensor}, signal processing~\citep{comon2009tensor} etc. 

In several / most of these applications, the primary tensor feature of interest is its low-rank factorization or approximation. Existing algorithms to compute the same, like alternating least squares~\citep{carroll1970analysis, harshman1970foundations}, tensor power method ~\citep{de2000best, anandkumar2014tensor} etc. need to access data in every iteration and can be computationally intensive. This is true both for settings where the tensor is already explicitly formed and available, and settings where it needs to be formed by taking appropriate outer products of data samples.

Methods involving tensors are intrinsically more computationally intensive as compared to, for example, those involving matrices; the focus of this paper is to provide a new and (at least a-priori) non-obvious technique to sample and compute sparse approximation of tensors. \\

\noindent {\bf Our contributions:} Our objective is to determine a small (random) subset of elements of a tensor that can be taken as a sparse surrogate for the tensor, in the sense that their spectral properties are similar. Our main contribution(s) is to develop a new way to determine this small subset in a data-dependent way, so that we can achieve this objective {\em without placing any incoherence-like assumptions on the underlying tensor}. We focus on three related but distinct settings for our three main contributions:
\begin{itemize}
\item {\it Direct tensor sparsification from samples:} This focuses on the common setting (especially in ML applications) where one is given samples $X_i \in \R^n$ and is interested in spectral properties of the outer product tensor $\mathcal{T} := \sum_{i=1}^p X_i \otimes X_i \otimes X_i$. We impose no additional structural assumptions on the tensor. Naively, this requires computing the $n^3$ elements of the tensor first, and random sampling can be quite bad. 

We instead provide a new (biased random) sampling distribution which allows us to choose as few as $m := O(\frac{n^{1.5} \log^3(n)}{\eps^2})$ elements to compute, yielding a sparse tensor $\widehat{\T}$ whose spectral error is bounded by $\lV \widehat{\T}- \T \rV \leq \eps \sqrt{n}*\left(\sum_{i=1}^p \|X^i\|^3 \right )$, w.h.p. Furthermore, our algorithm can compute the distribution with just one pass of all the samples (and thus requires two passes overall - the second one to actually compute the elements); the computational complexity is thus $O(nnz(X) +p*m*log(n))$. Sparse tensors are much easier to store and factorize.

%
%We develop a new algorithm that, with just one pass over all the samples, produces a sampling distribution for tensor elements. We show that if one computes as few as $O(\frac{n^{1.5} \log^3(n)}{\eps^2})$ elements of the $n^3$ total 
%
%The goal in tensor sparsification is to compute a sparse tensor that is a good approximation of a given tensor. Often the tensor is not given explicitly but is to be computed from tensor product of samples, $\sum_{i=1}^p X_i \otimes X_i \otimes X_i,~ X_i \in \R^n$. We present a new algorithm for sparsification/approximation for \emph{any} tensor without explicitly forming it. Existing techniques need all the tensor entries to be computed before sparsification. Our algorithm involves sampling few entries of a tensor and reweighing them according to a distribution computed \emph{only from the samples}.  
%
%This distribution can be computed efficiently in one pass from the samples and doesn\rq{}t need the computation of the tensor. We show that this generates a good approximation $\wT$ in spectral norm from few samples. Specifically, from $m \geq O(\frac{n^{1.5} \log^3(n)}{\eps^2})$ samples we guarantee error $\lV \widehat{\T}- \T \rV \leq \eps \sqrt{n}*\left(\sum_{i=1}^p \|X^i\|^3 \right)$, w.h.p. The computation complexity of the algorithm is $O(nnz(X) +p*m*log(n))$.

\item {\it Exact Tensor Completion:} In this setting, one wants to exactly recover a rank-$r$ orthogonal\footnote{Tensors that are not orthogonal have significantly harder algebraic structure.} tensor (i.e. $\T=\sum_{i=1}^r \so_i \Uo_i \otimes \Uo_i \otimes \Uo_i,~ \Uo_i \in \R^n$ are orthogonal ), from a small number of (randomly chosen) elements -- this represents the tensor generalization of the popular matrix completion setting. So far it is known that tensors with restrictive incoherence conditions can be recovered from a small number of uniformly randomly chosen elements~\citep{jain2014provable}. Incoherence however is a restrictive setting that precludes settings with high dynamic ranges (e.g. power laws) in element magnitudes.

We consider the case where the low-rank orthogonal tensor has no additional incoherence properties. We show that, if the samples come from a special distribution (that is ``adapted" to the underlying tensor) then the tensor can be provably exactly recovered from {\em (a)} as few as  $m= O( (\sum_{i=1}^n \|(\Uo)^i\|^{\ftt})^2 nr^3\kappa^4 \log^2(n) )$  samples w.h.p. {\em (b)} using a simple, fast and parallel weighted alternating least squares algorithm. The distribution depends only on the row norms of $\Uo$.  The algorithm has a low complexity of $O(mr^2)$, but performance depends on the restricted condition number $\kappa$.

\item {\it 2-pass Approximate Tensor Factorization:} Finally, we consider the problem -- again common in ML applications -- where we are given an arbitrary (large) low-rank orthogonal tensor that has been corrupted by arbitrary but bounded noise  ($\T =\sum_{i=1}^r \so_i \Uo_i \otimes \Uo_i \otimes \Uo_i + \E$), and we would like to approximately recover the orthogonal factors $\Uo$ fast. 

We provide a new algorithm to do so, that operates in two stages: {\em (a)} in the first stage, it takes one pass over all the elements of the tensor to determine a sampling distribution, and in a second pass extracts $ O( \frac{n^{1.5}}{\eps^2} r^3\kappa^4 \log^2(n) )$  elements of the tensor. Then {\em (b)} in the second stage, it uses these samples to do a tensor completion via weighted alternating least squares (which is fast, simple and parallel) to compute approximate factors with error $\|U_l -\Uo_l\| \leq \frac{12 \|\E\| }{\so_{\min}} +\eps\frac{\|\E\|_F }{\so_{\min}}$ with probability $\geq 1-\frac{1}{n^{10}}$.

%Further our algorithm needs only two passes over the data and is easily parallellizable.  

\end{itemize}

As mentioned, our algorithms needs only two passes over the data, are faster with less memory requirements and trivially parallellizable. Towards the end we will present some numerical simulations to illustrate our results. Note that we only discuss results for order-3 symmetric tensors for ease of notation. All our results extend to higher order non-symmetric tensors.\\

The rest of the paper is organized as follows: In Section~\ref{sec:related} we will first present some background of tensor factorization and later discuss related work. In Section~\ref{sec:tensor_spfcn} we will present our results on direct tensor sparsification from samples. In Section~\ref{sec:tensor_als} we will present our exact tensor completion results. In Section~\ref{sec:approx} we will discuss the 2-pass algorithm for computing factors of a tensor. Finally we present some results from numerical experiments in Section~\ref{sec:sims}.

\section{Background}\label{sec:related}

In this section we will present some background on tensor factorization and discuss related results. \\

\noindent {\bf Tensor Factorization:} An order-3 tensor $\T \in \R^{n \times n \times n}$, is of rank-$r$ if the minimum number of rank-1 tensors it can be decomposed is $r$, i.e., $\T =\sum_{l=1}^r u_l \otimes v_l \otimes w_l$, $u_l ,v_l$ and $w_l$ are vectors in $\R^n$. This decomposition is known as CANDECOMP/PARAFAC (CP) decomposition of a tensor. Rank of a tensor denotes the CP rank in the rest of this paper.

Note that unlike matrices, the components $u_l$ need not be all orthogonal. Surprisingly tensors have unique decomposition under simple conditions on Kruskal rank of the factors $u_l$~\citep{kolda2009tensor}. This makes tensor factorization appealing in latent variable learning in many applications like LDA, HMM, Gaussian mixture models, ICA ~\citep{anandkumar2014tensor, anandkumar2012spectral}. % and uniqueness is guaranteed under very simple conditions

In general finding factorization or even just the rank of a tensor is NP-hard~\citep{hillar2013most}. However if the tensor has orthogonal factorization then the factors can be computed using the tensor power method~\citep{de2000best, anandkumar2014tensor}. Recently \citep{anandkumar2014guaranteed} has given guarantees on factoring a tensor with incoherent (low inner product) factors. \citep{richard2014statistical} has analyzed various algorithms for recovering underlying factors from a spiked statistical model.  Note that these algorithms need to access the entire data over multiple iterations. %However their algorithms need access to the complete tensor.

In situations where tensor is computed as higher order moment from the samples, one can use the sample covariance matrix to perform whitening and convert the tensor factors into orthogonal factors~\citep{anandkumar2014tensor}. This also reduces the problem dimension and one can compute the factors fast using the tensor power method. However  this technique cannot be used in settings where one observes the entries of the tensor directly like ratings in a user*movie*time tensor, or EEG signal measuring electrical activity in brain as a time*spectral*space tensor. Our algorithm~\ref{algo:approx} in section~\ref{sec:approx}, computes factors fast by sampling few entries of the tensor and doing tensor completion.
%Simultaneous diagonalization.

Other popular factorization of a tensor is Tucker decomposition. Here we express tensor as a product of 3 orthogonal matrices $U\in \R^{n \times r_1}, V\in \R^{n \times r_2}, W\in \R^{n \times r_3}$ and a core matrix $A\in\R^{r_1 \times r_2 \times r_3}$, i.e., $\T_{ijk} =\sum_{pqr} U_{ip} V_{jq} W_{kr} A_{pqr}$. For a detailed discussion and algorithms we refer to~\citep{kolda2009tensor}.\\

\noindent {\bf Tensor Sparsification:} The goal in tensor sparsification problem is to compute a sparse sketch of a tensor. \citep{tsourakakis2010mach} has given a way to compute approximate factorization of a tensor but the approximation is in Frobenius norm. \citep{nguyen2010tensor} proposed a randomized sampling technique to compute sparse approximation. Specifically they sample entries with probability proportional to entry magnitude squared. They have given approximation guarantees in spectral norm with $O(\frac{ n^{1.5} \log^3(n)}{\eps^2})$ samples. We present similar guarantees in the setting where tensor is not already computed but one has access to the sample data (Section~\ref{sec:tensor_spfcn}).\\

\noindent {\bf Tensor Completion:} In tensor completion problem one wants to recover a low rank tensor from seeing only few entries of the tensor. There are multiple algorithms proposed for tensor completion without guarantees based on weighted least squares~\citep{acar2011scalable}, trace norm minimization~ \citep{liu2013tensor} and alternating least squares~\citep{walczak2001dealing}. \citep{ mu2014square, tomioka2011statistical} proposed various equivalents of nuclear norm for tensors and studied the problem of tensor completion but under Gaussian linear measurements, different from the setting considered here, where one sees the entries of the tensor. \\

%\citep{yuan2014tensor} has shown that using a tensor norm similar to nuclear norm of matrices, 
\citep{jain2014provable} has recently shown that one can recover a $\mu$-incoherent rank-$r$ orthogonal tensor from observing only $O(n^{1.5} \mu^6 r^5 \kappa^4 (\log n)^4 \log(r\|T\|_F/ \eps) )$ random entries. The authors use an alternating minimization style algorithm to recover the factors of the tensor. In Section~\ref{sec:tensor_als} we show exact recovery for any orthogonal tensor without incoherence assumption from fewer samples, if sampled appropriately.  Another recent work by~\citep{barak2015tensor} has given a sum-of-squares hierarchy based algorithm for prediction of incoherent tensors. However they only give approximation guarantees and not exact recovery. Interestingly their techniques work with general incoherent tensors and do not need orthogonality between factors.  \\

\noindent {\bf Matrix Completion:} In low rank matrix completion one wants to recover a low rank matrix from seeing only few entries of the matrix. This is a well studied problem starting with the work of~\citep{candes2009exact, candes2010power}  and later \citep{recht2009simpler, gross2011recovering} using nuclear norm minimization algorithm\footnote{Nuclear norm of a matrix is sum of its singular values.}. Other popular algorithms which guarantee exact recovery are OptSpace~\citep{keshavan2010matrix} and alternating minimization~\citep{jain2013low, hardt2013understanding}. These results assume that the underlying matrix is incoherent and the entries are sampled uniformly at random. \cite{chen2014coherent} has given guarantees for recovery of any rank-$r$, $n \times n$ matrix from $O(nr\log^2 (n))$ samples if sampled according to the leverage score distribution \footnote{Let SVD of $M$ be $U\Sigma V^T$, then $p_{ij} \propto \|U^i\|^2 + \|V^j\|^2$ is the leverage score distribution.}. \\

\noindent {\bf Matrix Approximation/Sparsification:} This is another active area with huge amount of interesting literature. Given a matrix $M$, the goal is to produce a low dimensional approximation (sketch) of the matrix with good approximation guarantees and in small number of passes over the data. The sketch can be a sparse matrix or a low rank matrix. Given the huge amount of literature we will not be able to do justice to all the works and we direct the interested reader to the nice survey articles~\citep{ halko2011finding, mahoney2011randomized, woodruff2014sketching}.   %~ \cite{drineas2011note, achlioptas2013near} or a low rank matrix.%~\citep{}.

Directly relevant to our 2-pass tensor factorization results (Section~\ref{sec:approx}) are the entrywise sampling results of~\citep{achlioptas2001fast, drineas2011note, achlioptas2013near, doi:10.1137/1.9781611973730.62} for matrices. In particular~\citep{achlioptas2001fast} proposed an entrywise sampling and quantization method for low rank approximation and has given additive error bounds. \citep{doi:10.1137/1.9781611973730.62}, has presented a low rank approximation algorithm using the leverage score sampling. Our work is similar in spirit to these results for matrices, but the techniques used for matrices like matrix Bernstein inequality do not extend to tensors and newer techniques are needed.\\

%Considers Gaussian linear measurements of tensor.. and nuclear norm equivalents~\citep{mu2014square, tomioka2011statistical}, but note that such methods based on convex optimization generally has higher computation complexity.\\

\noindent {\bf Notation}: 
Capital letter $U$ typically denotes a matrix and calligraphic letter $\T$ denote a tensor. $U^i$ denotes the $i$-th row of $U$, $U_j$ denotes the $j$-th column of $U$, and $U_{ij}$ denotes the $(i,j)$-th element of $U$. Unless specified otherwise, $U\in \R^{n\times r}$ and $\T \in \R^{n \times n \times n}$. $\T_{ijk}$ denotes the $(i, j, k)$ element of the tensor. $\|x\|$ denotes the L2 norm of a vector. $\|M\|=\max_{\|x\|=1}\|Mx\|$ denotes the spectral or operator norm of $M$. $\|M\|_F=\sqrt{\sum_{ij}M_{ij}^2}$ denotes the Frobenius norm of $M$. 

Now define tensor operation on a vector $\theta \in \R^n$ as follows, $$\T(I,\theta,\theta) =\sum_i(\sum_{jk}\T_{ijk} \theta_j \theta_k) e_i,$$ where $I$ is a $n \times n$ identity matrix. Spectral norm of a tensor $\T$ is defined as follows: $$ ||\T|| =\max_{u, v: ||u||, ||v||=1} \|\T(I,u,v)\|.$$ $\|T\|_F = \sqrt{\sum_{ijk} \T_{ijk}^2}$ denotes the Frobenius norm of $\T$.

$\Omega\subseteq [n]\times [n] \times [n]$ usually denotes the sampled set with, $P_{\Omega}(\T)$ is given by: $P_{\Omega}(\T)_{ijk}=\T_{ijk}$ if $(i,j,k)\in \Omega$ and $0$ otherwise. $R_{\Omega}(\T)=\W.*P_{\Omega}(\T)$ denotes the Hadamard product of $\W$ and $P_{\Omega}(\T)$.  Similarly let $R_{\Omega}^{1/2} (\T)_{ijk} =\sqrt{\W_{ijk}}\T_{ijk}$ if $(i,j,k)\in \Omega$ and $0$ otherwise.  $C$ is a constant independent of other parameters of the tensor and can change from line to line.

$\T_{i,:,:}$ denotes the $n \times n$ matrix with entry $(j, k)$ being $\T_{ijk}$. Finally $[r]$ denotes the set of integers form $1$ to $r$.\\

\section{Direct Tensor Sparsification from Samples}\label{sec:tensor_spfcn}

In this section we will present a new two pass algorithm for computing a sparse approximation of a tensor $\T=\sum_{i=1}^p X_i \otimes X_i \otimes X_i$, where $X_i$ are sample vectors in $\R^n$. Our algorithm involves first computing a specific distribution from $X$ and sampling entries of the tensor according to this distribution. Note that our algorithm will not need to form the complete tensor from the samples $\{ X_i\}$, but only compute few entries of the tensor that are sampled. Let $X$ be the sample matrix with $X_i$ as columns. Now we present the algorithm in detail.\\

\noindent {\bf Algorithm:} \{ {\it Input:} Data $X$ and sparsity $m$; {\it Output:} $O(m)$ sparse tensor $\Ro(\T)$.\}
\begin{itemize}
\item In one pass over the data $X$, compute $\|X^i\|,~ \forall i$.
\item Generate the sample set $\Omega$, where $(i, j, k) \in \Omega$ with probability $\widehat{p}_{ijk} = \min\{m*p_{ijk} ,1\}$, where \begin{equation}\label{eq:samp_prod}\pijk= \frac{\|X^i\|^3\|X^j\|^3 + \|X^j\|^3\|X^k\|^3 + \|X^k\|^3\|X^i\|^3}{3 n \|X\|_3^2}, \end{equation} $\|X\|_3=\sum_i \|X^i\|^3$. %and $m$ is the number of samples.
\item In one more pass over the data compute the tensor elements, $$\Ro(\T)_{ijk}= \frac{1}{\widehat{p}_{ijk}} \left(\sum_{l=1 }^p X_{il}X_{jl} X_{kl}\right),~ \forall (i, j, k) \in \Omega~ \text{and} ~0~ else.$$
\end{itemize} 

%We present a new distribution to do tensor sparsification. \begin{equation}\label{eq:samp_prod}\pijk= \frac{\|X^i\|^3\|X^j\|^3 + \|X^j\|^3\|X^k\|^3 + \|X^k\|^3\|X^i\|^3}{3 n \|X\|_3^2}, \end{equation}where $\|X\|_3=\sum_i \|X^i\|^3$. Let $m$ be the number of samples, then element $\T_{ijk}$ is sampled with probability $\widehat{p}_{ijk} = \min\{m*p_{ijk} ,1\}$. Define $\W_{ijk} =1/\widehat{p}_{ijk}$ when $\widehat{p}_{ijk} >0$, $0$ else. 

The output  of the algorithm is the sparse tensor $\Ro(\T)$, where $\Ro(\T)_{ijk} =\frac{\T_{ijk}}{\widehat{p}_{ijk}}$, if the entry is sampled and $0$ else. Now we will show that the sampled and reweighed tensor $\Ro(\T)$ is a good approximation to $\T$ in spectral norm.

\begin{theorem}\label{thm:tensor_init}
Given sample vectors $X_i \in \R^n$, let $\T=\sum_{i=1}^p X_i \otimes X_i \otimes X_i$. Then the sampled and reweighed tensor $\Ro(\T)$ generated according to the distribution~\eqref{eq:samp_prod}, satisfies the following:
\begin{equation}\label{eq:tensor_init}
\lV \Ro(\T) - \T \rV \leq \eps \left(\sqrt{n}*\sum_{i=1}^n \|X^i\|^3\right),
\end{equation}
with probability $\geq 1-\frac{1}{n^6}$, for $m \geq O(\frac{n^{1.5} \log^3(n)}{\eps^2})$.  %\frac{\|X\|_F}{\min_i \|X^i\|})$.
\end{theorem}

\noindent {\bf Remarks:}\\

\noindent  {\bf 1.} Expected number of sampled entries is $\leq m$. Hence the sparsity of the sampled tensor $\Ro(\T)$ is less than $2*m$ with high probability from concentration of binomial random variables.\\

\noindent  {\bf 2.} The proof of this theorem is discussed in appendix~\ref{sec:apdx_init} and relies mainly on appropriately partitioning the sets of $(i,j,k)$ and bounding error on each partition using the concentration bounds for spectral norm of a random tensor (Theorem~\ref{thm:tensor_conc} ~\citep{nguyen2010tensor}).\\
 
 %%Note that $\|T\|_F \leq \sqrt{p \sum_{l=1}^p \|X_l\|^6}.$

\noindent  {\bf 3.}  This theorem generalizes to an order-$d$ tensor $\T =\sum_{i=1}^p X_i \otimes^d$ with distribution $$p_{i_1, \cdots, i_d}= \frac{ \sum_{l \in [d]} \Pi_{q \in [d]/\l} \|X^q\|^{d/(d-1)} }{(d-1)n (\sum_{i=1}^n\|X^i\|^{d/(d-1)})^{d-1} }$$ and sample complexity $m \geq O(\frac{c^d n^{d/2}}{\eps^2}  \log^3(n) )$.\\ %where $n=\max_{i \in [d]} \{n_i\}.$\\

%\noindent {\bf Random Samples:} We compare the bound achieved in Theorem~\ref{thm:tensor_init} with $\|T\|_F$ for random tensors. 

\noindent  {\bf 4.}  
We now show that approximating the tensor in spectral norm gives constant approximation to the underlying factors if the tensor has orthogonal factors using the Robust Tensor Power Method (RTPM)~\citep{anandkumar2014tensor}. Intuitively good approximation is possible if the sample vectors do not cancel in adversarial way i.e., $\sqrt{n}*\left(\sum_{i=1}^p \|X^i\|^3 \right)$ is of the same order as $\so_{\min}$. Such an approximation to factors is desirable for initialization of algorithms like tensor power method or alternating least squares as we will discuss in the Section~\ref{sec:tensor_als}. The result follows from Theorem 5.1 of~\citep{anandkumar2014tensor}.

\begin{lemma}\label{lem:init}
Given a tensor $\T=\sum_{i=1}^p X_i \otimes X_i \otimes X_i$ with orthogonal factors s.t. $\T=\sum_{i=1}^r \so_i \Uo_i \otimes \Uo_i \otimes \Uo_i$ and $\eps \sqrt{n}*\left(\sum_{i=1}^p \|X^i\|^3 \right) \leq \frac{\so_{\min}}{r} $,  then, running $O(c \log r)$ iterations of RTPM on $\Ro(\T)$ sampled according to distribution~\eqref{eq:samp_prod}, gives $U_l$ and $\sigma_l$ satisfying the following.
 \begin{align}\label{eq:rtpm_init}
 \|U_l - \Uo_l\|_2 &\leq C  \eps \frac{\sqrt{n}*\left(\sum_{i=1}^p \|X^i\|^3 \right)}{\so_{l}} \\
 \lv \sigma_l -\so_l \rv &\leq  C  \eps \sqrt{n}*\left(\sum_{i=1}^p \|X^i\|^3 \right) ,
   \end{align}
   for all $l \in [r]$ with probability $\geq 1- \frac{1}{n^2}$, for $m \geq O(\frac{n^{1.5} \log^3(n)}{\eps^2})$ and constant $C$.
\end{lemma}

\noindent  {\bf  Computation complexity:} Computation of the sample distribution needs $O(nnz(X))$(sparsity of $X$) time. Computing $m$ entries of the tensor from the distribution has $O(p*m \log(n))$ time. Note that both these steps can be implemented in two passes over the data matrix $X$. On the contrary just computing the tensor from the samples takes $O(p n^3)$ time. Further one of the most performed operation with tensors, tensor-vector product $\Ro(\T)(I,v,v)$ takes only $O(m)$ time, independent of $p$, compared to $O(n*p)$ complexity of computing $(\T)(I,v,v)$.

%%In modern applications with growing data sizes computing the tensor or reading the tensor from the hard disk can be a huge bottleneck. There is interest in computing a sparse approximation to the tensor. In \citep{nguyen2010tensor} authors have presented a way to do tensor sparsification. But the distribution depends on the entries of $\T$ and computing the distribution itself can be a bottleneck from the samples.

%We now present the result for a orthogonal decomposable tensor. 
%\begin{corollary}
%For a orthogonal rank-$r$ tensor $\T =\sum_{l=1}^r \sigma_l U_l \otimes V_l \otimes W_l$, where $U, V, W$ are orthonormal matrices in $\R^{n \times r}$ and $\sigma_l \geq 0$. Let $\Ro(\T)$ be the sampled and reweighed tensor with the distribution \begin{multline*}\pijk= \frac{\|\T_{(i,:,:)}\|_F^3 \|\T_{(:,j,:)}\|_F^3 +  \|\T_{(:,j,:)}\|_F^3 \|\T_{(:,:,k)}\|_F^3 }{nz}\\+ \frac{\|\T_{(:,:,k)}\|_F^3 \|\T_{(i,:,:)}\|_F^3}{ n Z},\end{multline*} where $Z$ is the normalizing constant.
%\begin{equation*}
%\lV \Ro(\T) - \T \rV \leq \eps Z .
%\end{equation*}
%when $m \geq O(\frac{n^{1.5} \log^3(n)}{\eps^2})$. 
%\end{corollary}
\section{Exact Tensor Completion}\label{sec:tensor_als}

In this section we will present our main result on the tensor completion problem.  Let $\T =\sum_{i=1}^r \so_i \Uo_i \otimes \Uo_i \otimes \Uo_i$ where $\Uo \in \R^{n \times r}$ is an orthonormal matrix. The tensor completion problem is to recover the rank-$r$ tensor from observing only few entries. We will show that for any rank-$r$ tensor if entries are sampled according to a specific distribution, it can be recovered exactly from less than $O(n^{1.5 }r^3 \log^3(n))$ samples using algorithm~\ref{algo:wals}. \\%as few as $O(n^{1.5})$ samples.

%%Let $U \in \R^{n \times r}$ be the matrix with $\ul$ as columns. We assume the tensor is sampled from an underlying probability distribution. Once we have the sampled entries of the tensor we compute the factorization using the weighted alternating least squares algorithm~\ref{algo:wals}.

\noindent {\bf Sampling:} Now we will describe the sampling distribution that is sufficient to show exact recovery of any low rank orthogonal tensor from less than $O(n^{1.5} r^3 \log(n))$ samples. \begin{equation}\label{eq:tensor_ls} p_{ijk}= \frac{\| (\Uo)^i\|^{\ftt} \|(\Uo)^j\|^{\ftt}+ \|(\Uo)^j\|^{\ftt} \|(\Uo)^k\|^{\ftt} + \|(\Uo)^k\|^{\ftt} \|(\Uo)^i\|^{\ftt}}{3n (\sum_i \|(\Uo)^i\|^{\ftt})^2}.\end{equation} Let $m$ be the number of samples, then element $\T_{ijk}$ is sampled independently with probability greater than $\widehat{p}_{ijk} = \min\{m*p_{ijk} ,1\}$. The sampling distribution depends on the row norms of $\Uo$. We discuss the intuition for this distribution in Section~\ref{sec:disc}.\\

\noindent {\bf Algorithm:} For recovery of the tensor factors from the samples we use an alternating least squares algorithm~\ref{algo:wals}. Define the weights $\W_{ijk}=1/\widehat{p}_{ijk}$ when $\widehat{p}_{ijk} >0$, and $0$ else. The algorithm minimizes the error, $$\min_{U \in \R^{n \times r}} \sum_{ijk \in \Omega} \wijk \left(\T_{ijk}- \sum_{l=1}^r U_{il} U_{jl} U_{kl} \right)^2,$$ in an iterative way as discussed below. For detailed pseudocode look into algorithm~\ref{algo:wals}.\\

\noindent \{ {\it Input:} Sampled tensor $\Po(\T)$, Initialization-$U$, weights $\W$, iterations-$b$; {\it Output:} Completed tensor $\widehat{\T}$.\}
\begin{itemize}
\item Compute the sparse residual tensors, $\mathcal{R}_q = P_{\Omega}(\T -\sum_{l\neq q}\sigma_l U_l \otimes U_l \otimes U_l)$, for all $q \in [r]$.
\item Compute the updates $\wutp_q$ by solving the weighted least squares problems, $$ \wutp_q =\argmin_{u \in \R^n} \| R_{\Omega}^{1/2}\left(\mathcal{R}_q - u\otimes U_q \otimes U_q \right) \|_F^2,$$ for all $q \in [r]$.
\item Set $\sigma_q =\|\wutp_q\|$ and $U_q =\wutp_q/\sigma_q$, for all $q \in [r]$ and repeat the above steps for $b$ iterations.
\end{itemize}

%in each iteration by fixing two $U\rq{}s$ to the solution of the previous iterate and solving for the third $U$. By solving for one column of $U$ at a time this becomes just a weighted least squares problem. Note that this is fundamentally a non-convex problem, but we will show that the iterates converge to the global optima given sufficient number of samples.\\

%Now we will discuss how we minimize the loss function discussed above in detail. 

Note that this minimization is fundamentally a non-convex problem, but as we will see (Theorem~\ref{thm:tensor_als}) , the iterates converge to the global optima given sufficient number of samples.\\

\noindent Algorithm~\ref{algo:wals} needs good initialization with constant distance to the true factors as an input.  As we have seen in the previous section (Lemma~\ref{lem:init}), factors of the sampled and reweighed tensor $\Ro(\T)$ satisfy this condition for big enough number of samples $m$. Also we need to threshold each entry of $U$ at $2 \|(\Uo)^i\|$. We can estimate these values from the samples. \begin{equation}\label{eq:init} \|U_i -\Uo_i\| \leq \frac{1}{100r\kappa}, ~~\text{and} ~ ~ |U_{ij}|\leq 2\|(\Uo)^i\| ~~\forall i \in [r]. \end{equation} % as they are bounded random variables. %as shown in Lemma~\ref{lem:estim_thresh}.

Finally we assume that every iteration uses independent set of samples. This is to avoid dependence between successive iterates in the analysis. However this seems to be not required in practice as noticed in the simulations section. Now we will present the result about exact recovery of any orthogonal tensor using algorithm~\ref{algo:wals}.

\begin{theorem}\label{thm:tensor_als}
Let $\T =\sum_{i=1}^r \so_i \Uo_i \otimes \Uo_i \otimes \Uo_i$ be a rank-$r$ orthogonal tensor. Let $\Omega$ be generated according to~\eqref{eq:tensor_ls}. Then the output of algorithm~\ref{algo:wals} with initialization satisfying~\eqref{eq:init} after $b= O(4\sqrt{r}\log(\|\T\|_F/\eps))$ iterations satisfies the following:
\begin{equation}\label{eq:tensor_als}
\|T - \widehat{T} \| \leq \eps,
\end{equation}
with probability $\geq 1-\frac{1}{n^6}$, for number of samples $m \geq O((\sum_i \|(\Uo)^i\|^{\ftt})^2 nr^3\kappa^4  \log^2(n) \log(\|T\|_F/\eps) )$.
\end{theorem}

%%%%%%%%%% WALS algorithm%%%%%%%%%%%%%%%%%
%%%%%%%%%%%%%%%%%%%%%%%%%%%%%%%%%%%%%%%%%%
\begin{algorithm}[t]
\caption{WALS: Weighted Alternating Least Squares}
\label{algo:wals}
\begin{algorithmic}[1]
\INPUT $P_{\Omega}(\T),\  \Omega,\ r,\ P_{\Omega}(\W),\ b, \ U$
%\STATE $\W_{ijk}=1/\widehat{p}_{ijk}$ when $\widehat{p}_{ijk} >0$, $0$ else $\forall (i,j,k) \in \Omega$.
\STATE Divide $\Omega$ in $r*b$ equal random subsets, i.e., $\Omega=\{\Omega_1, \dots, \Omega_{r*b}\}$
%\STATE $R_{\Omega_0}(\T)\gets \W.*P_{\Omega_0}(\T)$
%\STATE $[U \gets [U^0_1, U^0_2, \cdots, U^0_r]$.
\FOR {$t=0$ to $b-1$}
	\FOR {$q=1$ to $r$}
		\STATE $\wutp_q= \argmin_{u \in \R^n}\| R_{\Omega_{t*r+q}}^{1/2}(\T- u \otimes U_q \otimes U_q -\sum_{l\neq q}\sigma_l U_l \otimes U_l \otimes U_l)\|_F^2$. 
		\STATE $\sigma^{t+1}_q= \|\wutp_q\| $.
		\STATE $\utp_q =\wutp_q/\|\wutp_q\|$.
  \ENDFOR
  \STATE $U \gets\utp$.
  \STATE $\Sigma \gets \Sigma^{t+1}$.
\ENDFOR
\OUTPUT Completed tensor $\widehat{T}= \sum_{l=1}^r \sigma_l U_l \otimes U_l \otimes U_l$. 
\end{algorithmic}
\end{algorithm}

%We will now discuss some implications of Theorem~\ref{thm:tensor_als}.\\
\noindent {\bf Remarks:}\\

\noindent {\bf 1.} Number of samples needed for exact recovery is $\leq 2m$ with high probability (from Binomial concentration). Hence we guarantee exact recovery of any rank-$r$ orthogonal tensor from $O( n  (\sum_i \|(\Uo)^i\|^{\ftt})^2 r^3\kappa^4 \log^2(n) )$ samples which is $\leq O(n^{1.5} r^{4.5}\kappa^4 \log^2(n) )$. This follows from $\sum_i \|(\Uo)^i\|^{\ftt} \leq  r^{3/4} n^{1/4}$. So for tensors with biased factors where $\sum_i \|(\Uo)^i\|^{\ftt}$ is a constant, need only $O(n)$ samples for exact recovery. The worst case is when the factors are incoherent and our sample complexity $O(n^{1.5})$ matches that of~\citep{jain2014provable}. This is the first such result to guarantee exact recovery of arbitrary orthogonal tensor and characterize the sample complexity for higher order tensors as far as we know.\\

\noindent {\bf 2.} The theorem generalizes to an order-$d$ tensor $\T \in \R^{n \otimes^d}$ with distribution $$p_{i_1, \cdots, i_d}= \frac{ \sum_{l \in [d]} \Pi_{q \in [d]/\l} \|(\Uo)^{i_q}\|^{d/(d-1)}}{dn (\sum_{i=1}^n\|(\Uo)^i\|^{d/(d-1)})^{(d-1)}}$$ and sample complexity $m \geq O((\sum_{i=1}^n\|(\Uo)^i\|^{d/(d-1)})^{(d-1)} nr^3 \kappa^4 \log^2(n) \log(\|T\|_F/\eps))$.\\%, where $n=\max_{i \in [d]} \{n_i\}.$\\

\noindent {\bf 3.} Algorithm~\ref{algo:wals} maintains only the factors of the tensor and the samples in each iteration. So it needs only $O(n* r +m)$ memory. Further since each iteration involves solving a weighted least squares problem, the computation complexity of the algorithm is $O(mr + m+ n)r*\log(\|\T\|_F/\eps)$, which is just $O(mr^2\log(\|\T\|_F/\eps))$. Hence this algorithm has low computation complexity and further each iteration can be easily parallelized.\\

\noindent {\bf 4.} The proof of Theorem~\ref{thm:tensor_als} similar to the proof technique of~\citep{jain2014provable}, involves showing a distance of the factors in the current iterate to the optimum decreases in each iteration (Lemma~\ref{thm:infty}). However our sampling distribution is not exactly uniform and the underlying tensor is not incoherent, so we have to carefully use the properties of the distribution~\eqref{eq:tensor_ls} to show convergence for arbitrary factors. The complete proof is presented in Section~\ref{sec:app_2}.

%\noindent {\bf 4.} \citep{chen2014coherent} has proved a similar result for matrices using leverage score sampling. This is the first such result to describe the distribution and characterize the sample complexity for higher order tensors as far as we know.\\

\subsection{Discussion}\label{sec:disc} Now we will discuss the intuition for the sampling~\eqref{eq:tensor_ls}. The distribution~\eqref{eq:tensor_ls} is important to guarantee exact recovery. The key idea is, distributions like $L1$, $L2$ and ~\eqref{eq:sum_l3} do not sample enough entries corresponding to biased factors and some distributions like~\eqref{eq:prod_l3} do not sample enough entries corresponding to the unbiased factors. Proposed distribution~\eqref{eq:tensor_ls} achieves the right balance.% between sampling blocks of all sizes.

%bias towards the small blocks in the tensor enough and some distributions like~\eqref{eq:prod_l3} bias too much towards the smaller blocks and not enough on the larger blocks. Proposed distribution~\eqref{eq:tensor_ls} achieves the right balance between sampling blocks of all sizes.

Clearly with uniform distribution one cannot guarantee exact recovery unless one samples all the entries (for example consider rank-1 tensor which have single non zero entry). Now consider data dependent distributions where probability of sampling an entry is proportional to magnitude of the entry ($L1$) or magnitude squared ($L2$) of the tensor.  Now we will present a counter example for these distributions. %We will refer to the distribution~\eqref{eq:tensor_ls} as tensor leverage sampling in this section.

\begin{claim}\label{claim1}
There exists a rank-2 tensor for which sampling with $L1$ or $L2$ distributions, error is bounded away from zero, for number of samples $m \leq \frac{n^3}{\log^3(n)}$, w.h.p.
\end{claim}

\begin{proof}
Consider a rank-2 block diagonal tensor with the first block of size $\log^3 (n)$ of all ones and the second block of size $(n-\log (n))^3$ of all ones. The factors of this tensor are, $$u_1 =[\frac{1}{\sqrt{\log (n)}}, \underbrace{\cdots}_{\log (n)-2 }, \frac{1}{\sqrt{\log (n)}}, 0, \cdots, 0  ]^T ~~\text{and}~~ u_2=[0, \underbrace{\cdots}_{\log (n)-2 }, 0, \frac{1}{\sqrt{n - \log(n)}}, \cdots,\frac{1}{\sqrt{n - \log(n)}} ]^T.$$

Now with $L1$ sampling expected number of entries seen in the first block is $\approx m* \frac{\log^3 (n)}{n^3}$ which is less than 1 for $m \leq \frac{n^3}{\log^3(n)}$. Similarly $L2$ sampling also fails to sample the first block. Hence the error is bounded away from zero.

For the proposed sampling~\eqref{eq:tensor_ls} expected number of entries sampled in the first block is $\approx m* \frac{1}{n^{1.5} \log^{1.5} (n)}$. Hence the complete block is sampled for $m \geq O(n^{1.5} \log^{4.5} (n))$.
\end{proof}

Now consider more biased distributions. \begin{equation}\label{eq:sum_l3} p_{ijk}= \frac{\| (\Uo)^i\|^{3} + \|(\Uo)^j\|^{3} + \|(\Uo)^k\|^{3} }{3n^2 (\sum_i \|(\Uo)^i\|^{3})}.\end{equation}
and
\begin{equation}\label{eq:prod_l3} p_{ijk}= \frac{\| (\Uo)^i\|^{3} \|(\Uo)^j\|^{3}+ \|(\Uo)^j\|^{3} \|(\Uo)^k\|^{3} + \|(\Uo)^k\|^{3} \|(\Uo)^i\|^{3}}{3n (\sum_i \|(\Uo)^i\|^{3})^2}.\end{equation}

\begin{claim}\label{claim2}
There exists a rank-2 tensor for which sampling with distributions~\eqref{eq:sum_l3} or~\eqref{eq:prod_l3} , error is bounded away from zero, for number of samples $m \leq n^2/\log^2(n)$, w.h.p.
\end{claim}

\begin{proof}
Consider the same example as in Claim~\ref{claim1}, the rank-2 block diagonal tensor with the first block of size $\log^3 (n)$ of all ones and the second block of size $(n-\log (n))^3$ of all ones. The factors of this tensor are, $$u_1 =[\frac{1}{\sqrt{\log (n)}}, \underbrace{\cdots}_{\log (n)-2 }, \frac{1}{\sqrt{\log (n)}}, 0, \cdots, 0  ]^T ~~\text{and}~~ u_2=[0, \underbrace{\cdots}_{\log (n)-2 }, 0, \frac{1}{\sqrt{n - \log(n)}}, \cdots,\frac{1}{\sqrt{n - \log(n)}} ]^T.$$

Now with distribution~\eqref{eq:sum_l3}, expected number of entries sampled in the first block is $\approx m* \frac{1}{n^{2.25} \log^{1.5} (n)}$. Hence for $m \leq n^2$ first block is not sampled w.h.p. and the error is bounded away from zero.

Now consider the distribution~\eqref{eq:prod_l3}, expected number of entries sampled in the second block is $\approx m* \frac{\log^{0.5} (n) }{n}$. Hence for $m < n^2/\log^2(n)$, number of entries sampled is strictly less than $n-\log(n)$. Since second block has $n-\log(n)$ faces, atleast one face of the tensor is not sampled along each dimension and hence cannot be recovered.

%For tensor leverage sampling expected number of entries sampled in the first block is $\approx m* \frac{1}{n^{1.5} \log^{1.5} (n)}$. So one will see the sample the complete block for $m \geq O(n^{1.5} \log^{4.5} (n))$.
\end{proof}

%%%%%%%%%%%%%%%%%%%%%%%%%%%%%%%%%%%%%%
%%%%%%%%%%%%%%%%%%%%%%%%%%%%%%%%%%%%%%
\section{2-pass Approximate Tensor Factorization}\label{sec:approx}

In this section we will present a new algorithm to compute factors of an orthogonal tensor corrupted by noise. Our algorithm needs only two passes over the data unlike the existing algorithms which need to access the data over multiple iterations. Let $\T = \sum_{l=1}^r \so_l \Uo_l \otimes^3 + \E$, where $\Uo \in \R^{n \times r}$ is an orthonormal matrix and $\E$ is arbitrary but bounded noise. We specifically make the following assumptions on the noise. 
\begin{equation}\label{eq:noise_assume}
\|\E\|_F \leq C\frac{\so_{\min}}{100*r} ~~\text{and}~~ \|\E\|_{\infty} \leq \frac{\|\E\|_F }{n^{1.5}},
\end{equation}
where $\|\E\|_{\infty}$ is $\max_{ijk} \lv \E_{ijk} \rv$.

Now we will describe the algorithm we use to compute the factors of the tensor $\T$. The algorithm consists of two parts. First we sample the entries of $\T$ according to a specific biased distribution and then use algorithm~\ref{algo:wals} with the sampled entries to compute the factors.\\

\noindent {\bf Sampling:} Now we will describe the distribution used to sample the tensor. Note that unlike in previous section, the tensor is not exactly low rank and so the distribution is modified to account for the noise. Consider the following distribution which can be computed easily in one pass over the tensor.

\begin{align}\label{eq:samp_approx}
\pijk = 0.5\frac{\nu_i^{\ftt} \nu_j^{\ftt} +\nu_j^{\ftt} \nu_k^{\ftt} + \nu_k^{\ftt} \nu_i^{\ftt}}{3nZ}
+0.5 \frac{\T_{ijk}^2}{\|\T\|_F^2},
\end{align}
where $\nu_i = \frac{\|\T_{i, :, :}\|_F}{\|\T\|_F} + \frac{1}{\sqrt{n}}$ and $Z= \left(\sum_{i=1}^n \nu_i^{\ftt} \right)^2 $ is the normalizing constant. $\|T_{i, :, :}\|_F $ is the Frobenius norm of the $i$th face of the tensor and $\|\T_{i, :, :}\|_F^2 = \sum_{jk} \T_{ijk}^2$. Note that we use $\frac{\|T_{i, :, :}\|_F}{\|\T\|_F} $ as an estimate for $\|(\Uo)^i\|$. 

We compute factors $U$ of the sampled tensor $\Ro(\T)$ using RTPM and use them for initialization for the second step of the algorithm. Note that we also threshold the factors such that $U_{il} \leq 2 \nu_i$. \\

\noindent {\bf WALS:} The second part of the algorithm uses the samples from the first part and computes the factors using the WALS algorithm~\ref{algo:wals}. The intuition is, if the underlying tensor is exactly rank-$r$, then this reduces to the completion setting discussed in the previous section and algorithm~\ref{algo:wals} will indeed recover the underlying factors. Since the tensor is not exactly rank-$r$ it will introduce an error in the recovered factors.\\

The pseudocode of the algorithm is given in~\ref{algo:approx}. Now we will present the main recovery result.

\begin{theorem}\label{thm:approx}
Given a tensor $\T = \sum_{l=1}^r \so_l \Uo_l \otimes^3 + \E$, where $\Uo \in \R^{n \times r}$ is an orthonormal matrix and $\E$ satisfies assumption \eqref{eq:noise_assume}, the output of Algorithm~\ref{algo:approx} satisfies the following:
\begin{align*}
\|U_l -\Uo_l\| &\leq\frac{12 \|\E\| }{\so_{\min}} +\eps\frac{\|\E\|_F }{\so_{\min}} * \frac{\sqrt{Z}}{n^{0.25}} , ~ \forall l \in [r]~~\text{and}\\
\|\widehat{\T} -T\| &\leq 48r\kappa \|\E\| + \eps \|\E\|_F*\frac{\sqrt{Z}}{n^{0.25}},
\end{align*}
with probability $\geq 1-\frac{1}{n^{10}}$,~ for $m \geq O(\frac{n^{1.5}}{\eps^2} r^3 \kappa^4  \log^2(n) \log(4 \sqrt{r}\|T\|_F/\eps \|\E\|_F))$.
\end{theorem}
\noindent {\bf Remarks:}\\

\noindent {\bf 1.} Note that $Z= \left(\sum_{i=1}^n \nu_i^{\ftt} \right)^2 \leq 4n^{0.5}$. Hence $\frac{\sqrt{Z}}{n^{0.25}} \leq 2$. So for a tensor concentrated in few entries, $Z$ can be as small as a constant and hence the error is smaller for such tensors.  \\

\noindent {\bf 2.} In the error expression above, the first term $O(\|\E\|)$, arises even for algorithms that access the complete tensor~\citep{anandkumar2014tensor, anandkumar2014guaranteed}. The second term in the error $O(\eps \|\E\|_F)$, is the approximation error and decreases with increasing number of samples ($m$).\\

\noindent {\bf 3.} The assumptions on the noise~\eqref{eq:noise_assume} are satisfied by entrywise random Gaussian noise. Let $\E$ be a random tensor with $\E_{ijk} \sim \mathcal{N} (0, \frac{1}{n^{1.5}}) * \frac{\so_{\min}}{C\log(n)}$. Then $\|\E\|_{\infty} \leq \frac{C\so_{\min}}{n^{1.5}}$ and $\|\E\|_F \leq C\so_{\min}$ with high probability and $\E$ satisfies~\eqref{eq:noise_assume}.\\

%\noindent {\bf 3.} The assumptions on noise in the above theorem are stronger compared to the robust tensor power method that uses all the entries of the tensor to compute the factors. This is because the sampling step introduces additional noise. Thus this is a tradeoff between the computational complexity and the error.\\ %Hence the algorithm recovers the factors with an error proportional to noise Frobenius norm. .\\

%\noindent {\bf 3.} Thresholding step

%\noindent {\bf 2.} The algorithm 

\noindent {\bf Computation and memory:} Algorithm~\ref{algo:approx} has a complexity of $O(nnz(\T)+ mr^2)$ as the sampling step takes $O(nnz(\T))$ time and the algorithm~\ref{algo:wals} takes $O(mr^2)$ time. Hence by Theorem~\ref{thm:approx}, the complexity becomes $O(nnz(\T) + O(\frac{n^{1.5}}{\eps^2} r^5 \kappa^4  \log^2(n) \log(\|T\|_F/\eps))$. Further the sampling part of algorithm needs to read data and store only $O(n)$ numbers corresponding to distribution~\eqref{eq:samp_approx} and the WALS step needs only $O(m + n*r)$ memory in each iteration.

%%%%%%%%%% Approx algorithm%%%%%%%%%%%%%%%%%
%%%%%%%%%%%%%%%%%%%%%%%%%%%%%%%%%%%%%%%%%%
\begin{algorithm}[t]
\caption{Approximate tensor factorization}
\label{algo:approx}
\begin{algorithmic}[1]
\INPUT Tensor $\T$, number of samples $m$, rank $r$, iterations $b$.
\STATE In one pass over the data compute $\|T_{i, :, :}\|_F, \forall i$.
\STATE Compute samples $\Po(\T)$ from the tensor according to distribution~\eqref{eq:samp_approx} in one pass over the data.
\STATE Compute factors $U$ using robust tensor power method from $\Ro(\T)$.
\STATE Threshold $U_{ij}$ at $2\nu_i, \forall (i,j).$
\STATE $\{ U_i, \sigma_i \}_{i \in [r]}$ =WALS($\Po(\T), \Omega, r, \Po(1/\widehat{p}), b ,U$).
\OUTPUT $\{ U_i, \sigma_i \}_{i \in [r]}$.
\end{algorithmic}
\end{algorithm}

%\input{tensor_approx}
%\input{tensor_discuss}
%\documentclass[10pt]{article}
%\usepackage{natbib}
%\usepackage{graphicx}
%\usepackage{color}
%\usepackage{epstopdf}
%\def\T{\mathcal{T}}
%\begin{document}

%Simulations on sparsifying tensor (tensor not present).. err approx spectral norm

%simulations on exact recovery~diff dists

%simulations on noisy approx.

%\begin{figure*}[t!]
%    \centering
%    \begin{subfigure}[t]{width=\columnwidth}
%        \centering
%        \includegraphics[height=1.2in]{\label{fig:a}\includegraphics{figures/approx50}}
%        %\caption{Lorem ipsum}
%    \end{subfigure}%
%    ~ 
%    \begin{subfigure}[t]{width=\columnwidth}
%        \centering
%        \includegraphics[height=1.2in]{\label{fig:b}\includegraphics{figures/approx50}}
%        %\caption{Lorem ipsum, lorem ipsum,Lorem ipsum, lorem ipsum,Lorem ipsum}
%    \end{subfigure}
%    \caption{Caption place holder}
%\end{figure*}

%\begin{figure}
%\centering     %%% not \center
%\subfigure[Figure A]{\label{fig:a}\includegraphics[width=60mm]{figures/approx50}}
%\subfigure[Figure B]{\label{fig:b}\includegraphics[width=60mm]{figures/approx50}}
%\caption{my caption}
%\end{figure}
\section{Simulations}\label{sec:sims}

\begin{figure*}[h]
%\vskip -0.2in
  \centering
  \begin{tabular}[ht]{ccc}
    \includegraphics[width=.5\textwidth]{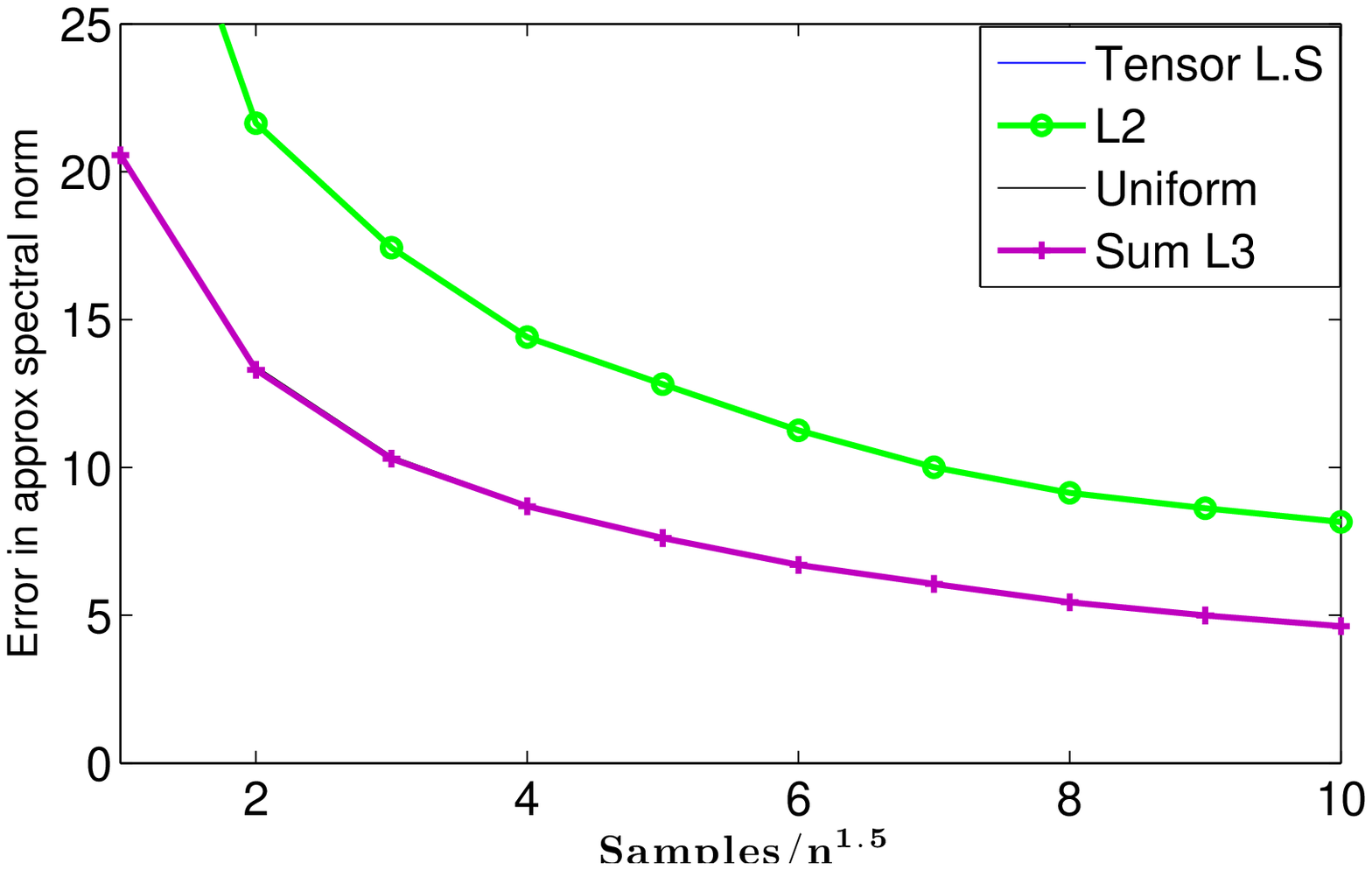}\hspace*{-0pt}&\includegraphics[width=.5\textwidth]{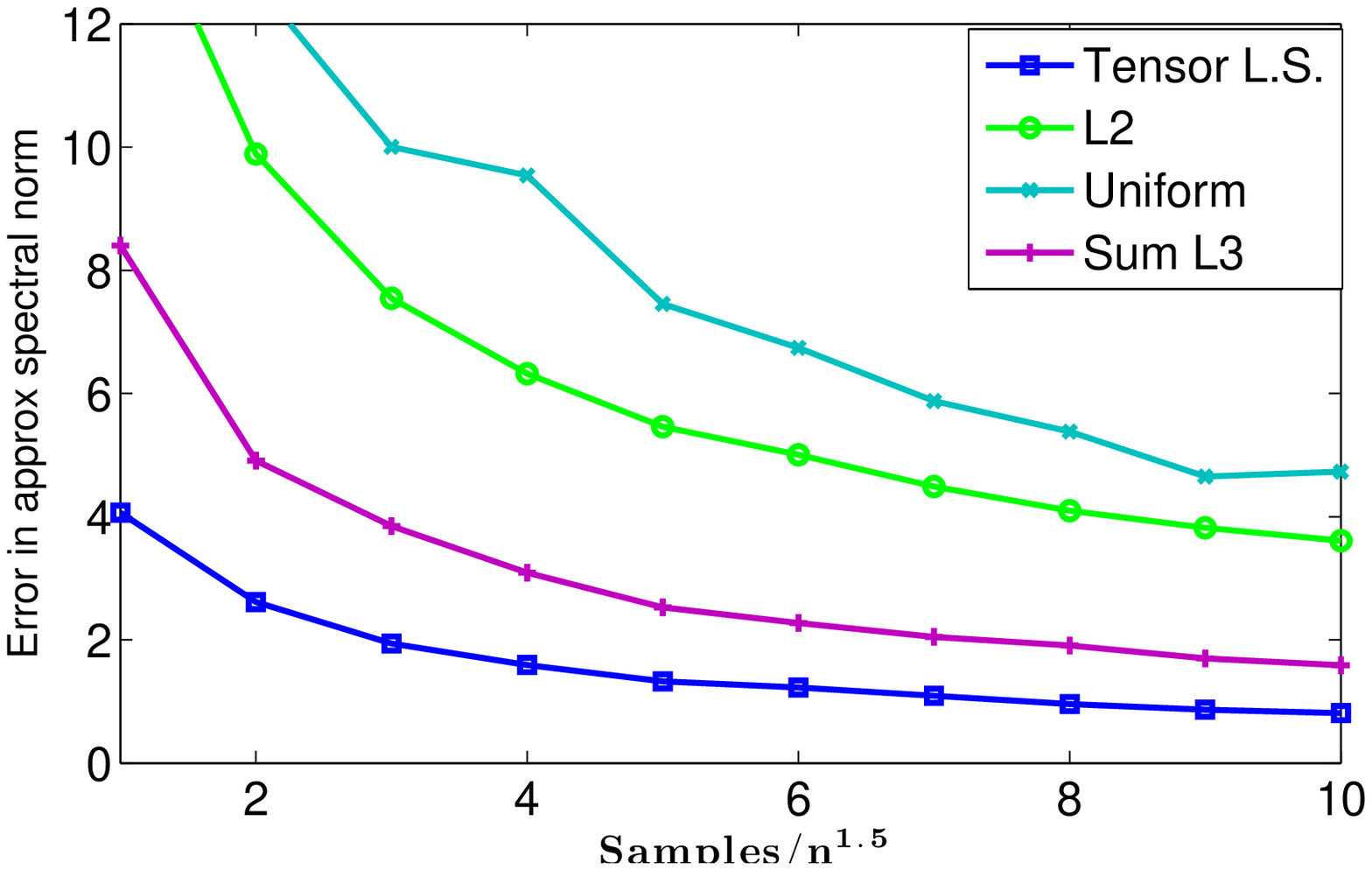}\hspace*{-15pt}\\
{\bf (a)}&{\bf (b)}
  \end{tabular}
  \caption{In this plot we compare error of various sampling distributions used to sample a random tensor $\T= \sum_{i=1}^p X_i \otimes X_i \otimes X_i$, as we increase the number of sampled entries. Notice that since we cannot compute the spectral norm of the error tensor we compute $L_{2,2}$ norm of the error. {\bf(a)}:  In the first plot we consider a tensor formed from random vectors $X_i$. For such tensors we notice that most sampling distributions including uniform work well.  {\bf(b)}: In this plot we create tensor from biased factors $D*X_i$, where $D$ is a diagonal matrix $D_{ii} = \frac{1}{i^a}$ with $a=0.5$. In this case we notice that the proposed sampling distribution achieves smaller error compared to other distributions.}
  \label{fig:plot12}
\vskip -0.1in
\end{figure*} 

\begin{figure*}[h]
\vskip -0.2in
  \centering
  \begin{tabular}[h]{ccc}
    \includegraphics[width=.5\textwidth]{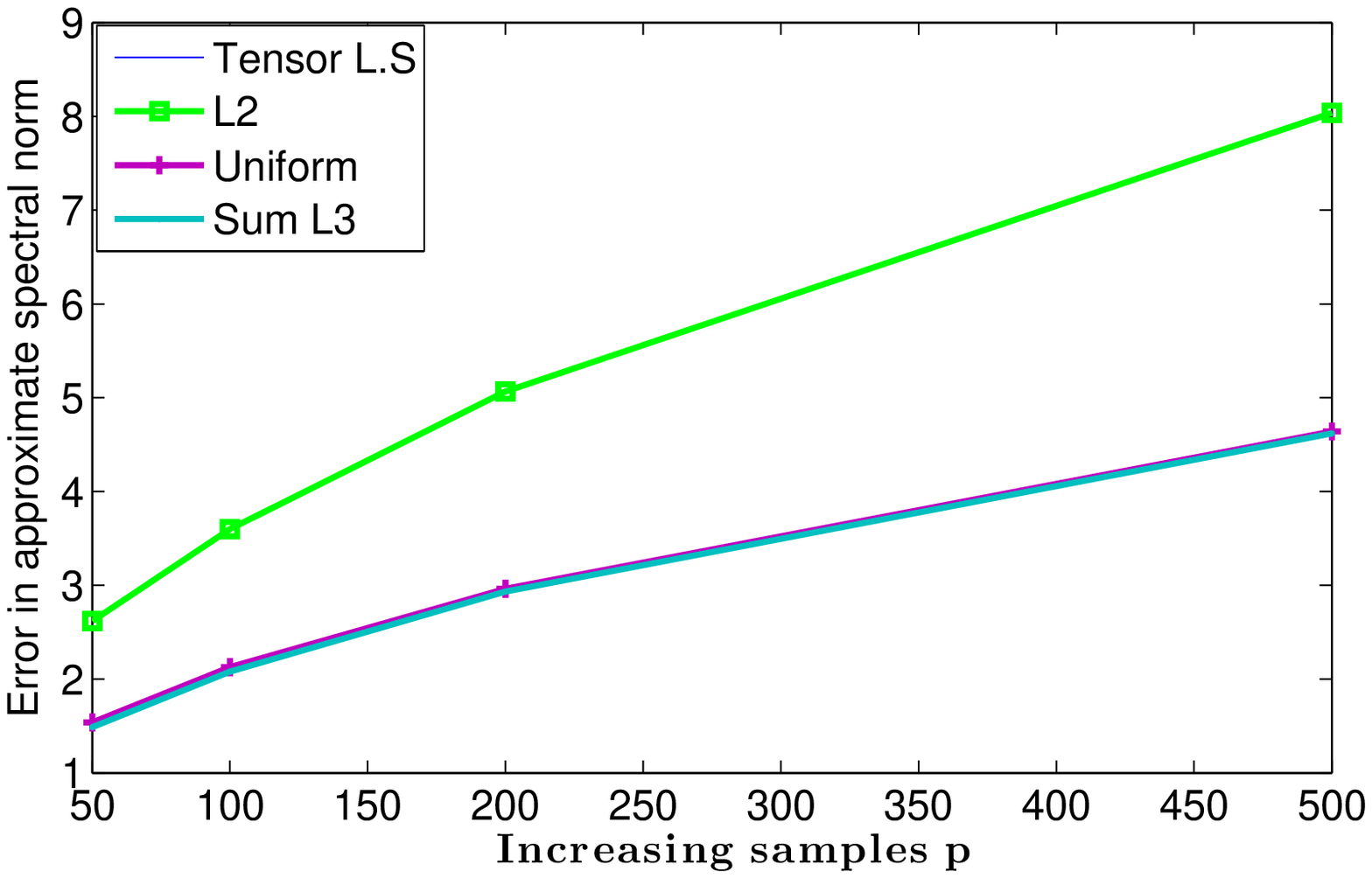}\hspace*{-0pt}&\includegraphics[width=.5\textwidth]{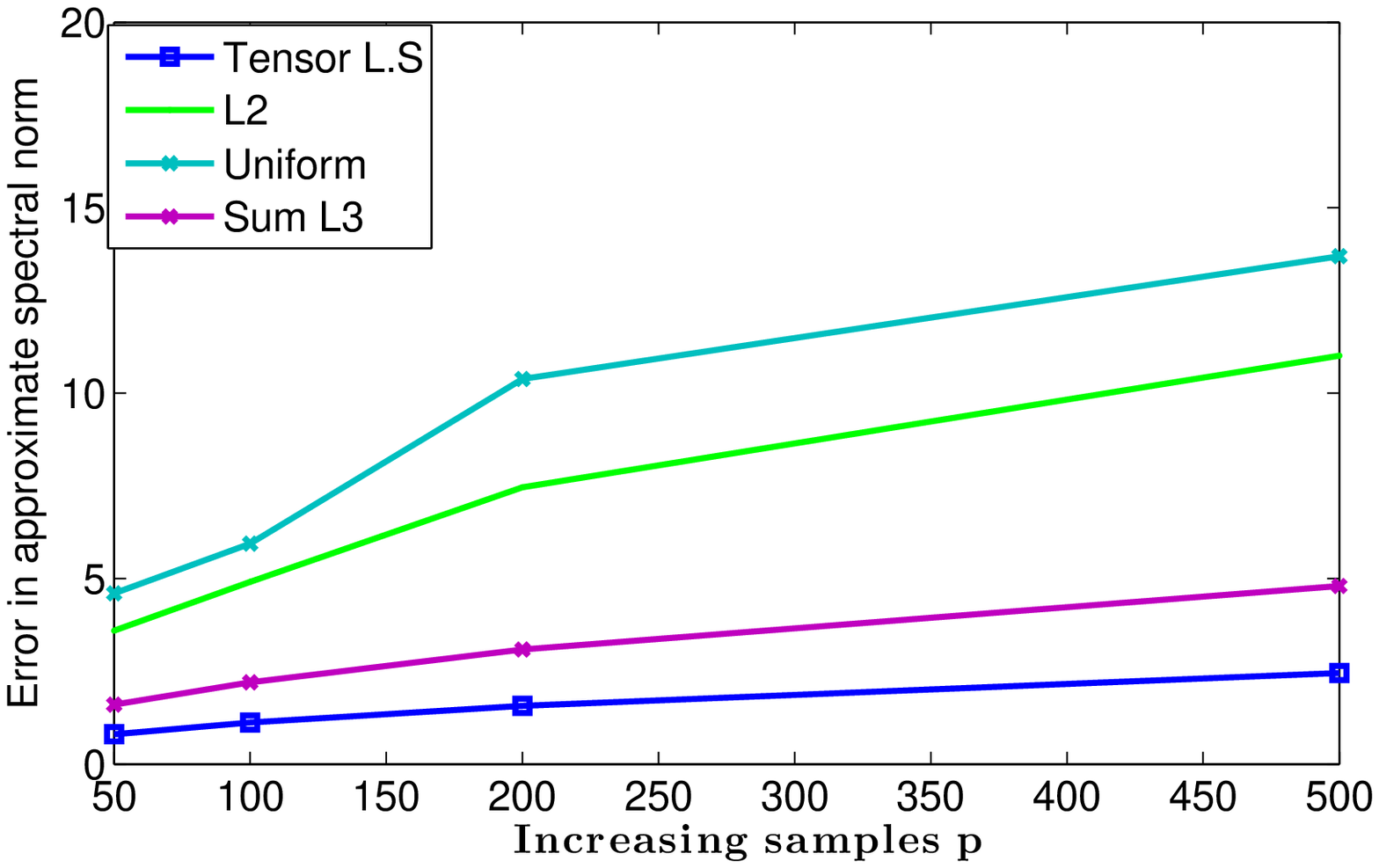}\hspace*{-15pt}\\
{\bf (a)}&{\bf (b)}
  \end{tabular}
  \caption{ In this plot we compare error performance of various sampling distributions, used to sample a random tensor, as we increase the number of sample vectors $p$. Note that as we increase the sample vectors $p$ the approximation becomes bad and the error increases.  {\bf(a)}: In the first plot we again consider a random tensor $\T= \sum_{i=1}^p X_i \otimes X_i \otimes X_i$, and most sampling distributions including uniform have similar error. {\bf(b)}: In this plot again we create a tensor from biased factors. In this case we notice that the proposed sampling distribution achieves smaller error compared to other distributions. }
  \label{fig:plot34}
\vskip -0.2in
\end{figure*} 

\begin{figure*}[h]
\vskip -0.2in
  \centering
  \begin{tabular}[h]{ccc}
    \includegraphics[width=.5\textwidth]{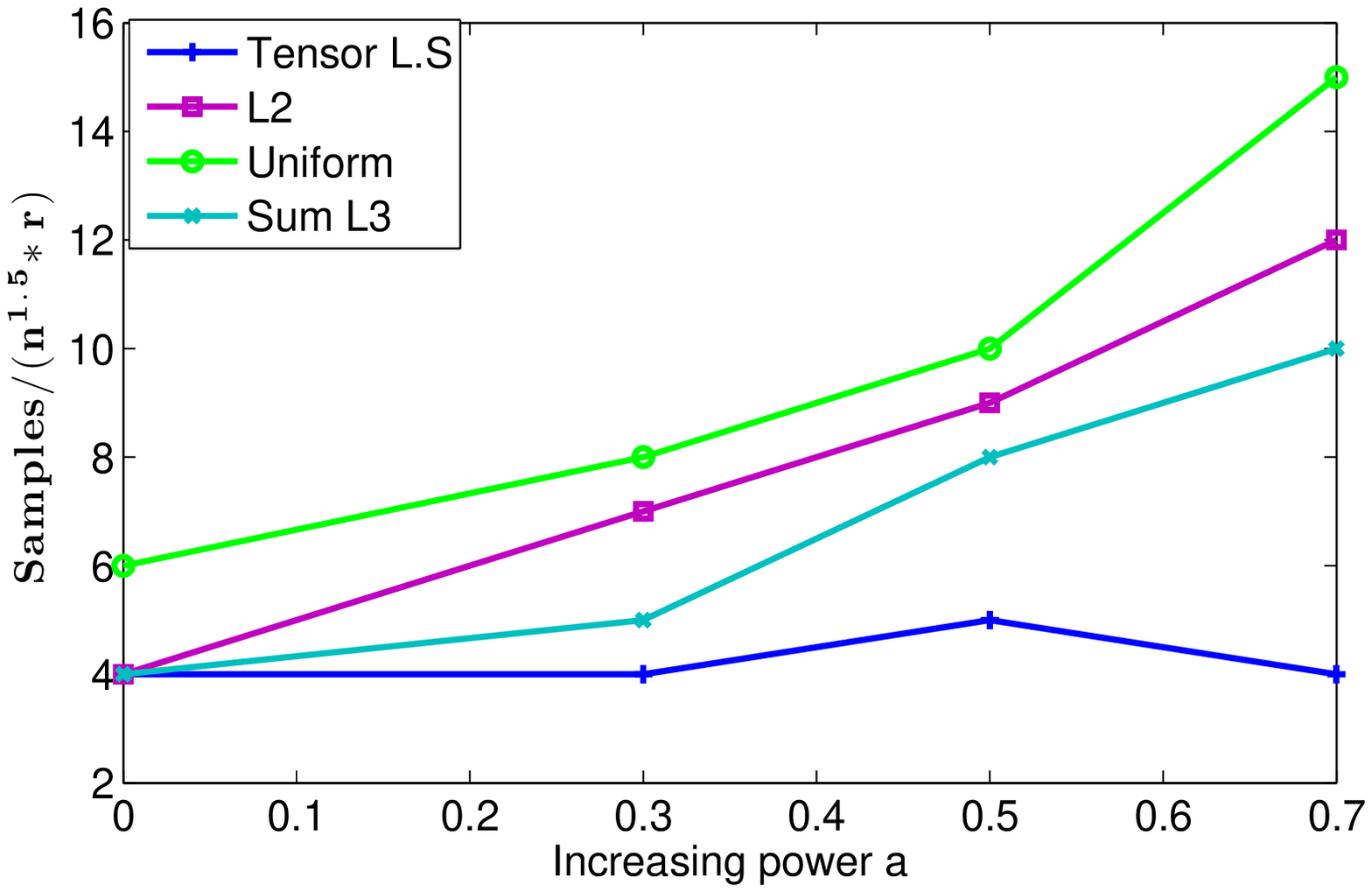}\hspace*{-0pt}&\includegraphics[width=.5\textwidth]{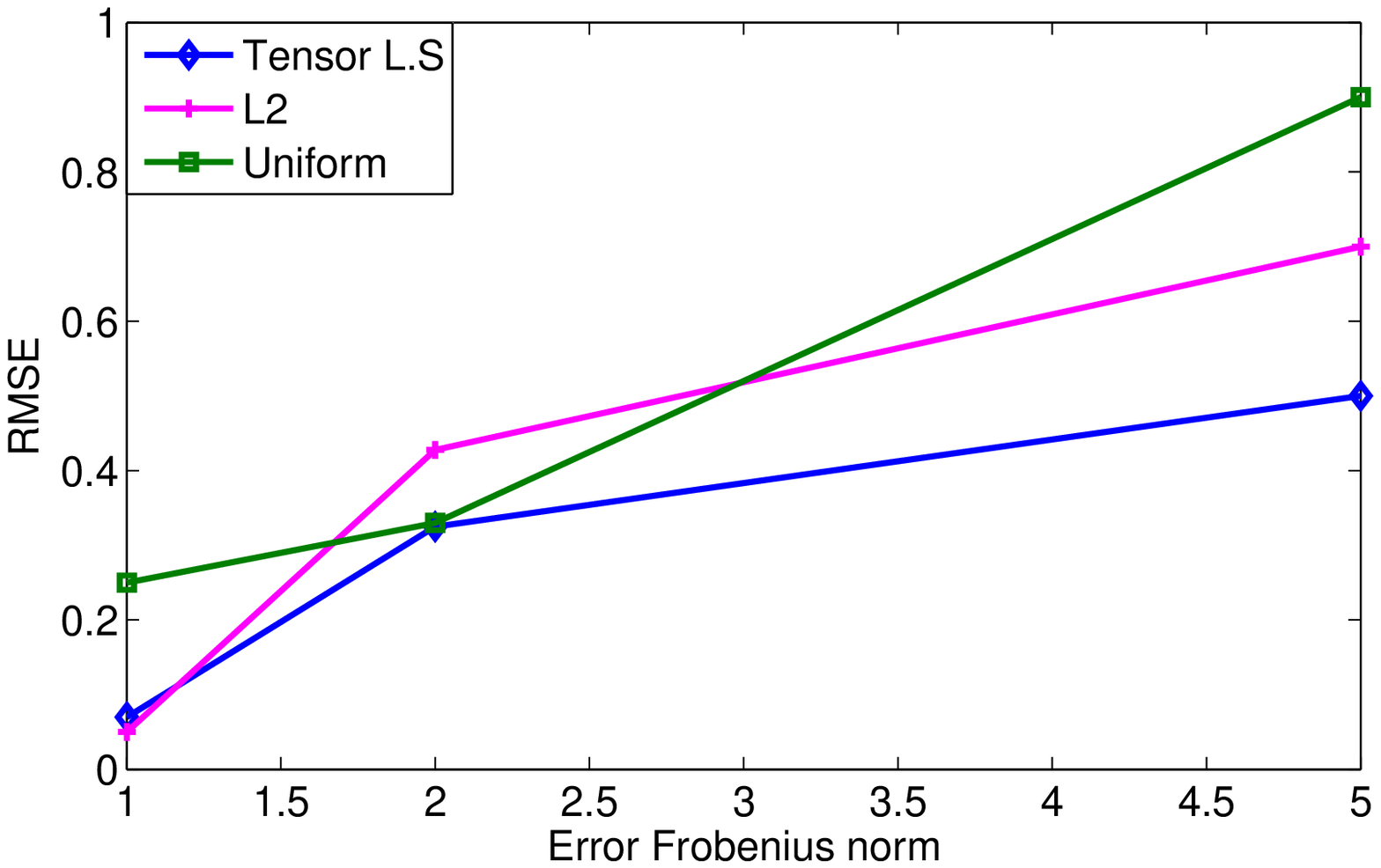}\hspace*{-15pt}\\
{\bf (a)}&{\bf (b)}
  \end{tabular}
  \caption{{\bf(a)}: In this plot we compare the number of samples needed for exactly recovering a rank-5 orthogonal tensor from different sampling distributions using algorithm~\ref{algo:wals}. $\T= \sum_{i=1}^5 U_i \otimes U_i \otimes U_i$, $U_i$ are orthogonal biased vectors with $U =SVD(D*X)$, where $X$ is a random matrix and $D$ is a diagonal matrix with $D_{ii} =\frac{1}{i^{a} }$. With increasing values of $a$ (x-axis) the tensor becomes concentrated on fewer entries. On y-axis we plot the number of samples needed for successful recovery (RMSE $< 0.01$) in more than $80\%$ runs. The proposed sampling distribution tensor L.S is able to recover the tensor from smaller number of entries even if the tensor gets biased. {\bf(b)}:  In this plot we consider the performance of algorithm~\ref{algo:approx} in the noisy tensor case $\T= \sum_{i=1}^5 U_i \otimes U_i \otimes U_i +\E$. $\E$ is an entry-wise random tensor. We plot the RMSE of the computed factors from algorithm~\ref{algo:approx} as the noise Frobenius norm increases. We notice that the proposed sampling distribution has smaller error.}
  \label{fig:plot56}
\vskip -0.2in
\end{figure*} 

%\begin{figure*}[ht]
%\vskip 0.2in
%\begin{center}
%\centerline{\includegraphics[width=120mm]{figures/completion}}
%\caption{  In this plot we compare number of samples needed for exactly recovering a rank-5 orthogonal tensor from different sampling distributions using algorithm~\ref{algo:wals}. $\T= \sum_{i=1}^5 U_i \otimes U_i \otimes U_i$, $U_i$ are orthogonal biased vectors with ($U =SVD(D*X)$, where $X$ is a random matrix and $D$ is a diagonal matrix with $D_{ii} =\frac{1}{i^{a} }$. As we go right on x-axis the value of $a$ increases making the tensor concentrated on fewer entries. On y-axis we plot number of samples needed for successful recovery (rmse < 0.01) in more than $80\%$ runs. The proposed sampling distribution tensor L.S is able to recover the tensor from the same number of entries even if the tensor gets biased. Other sampling distributions need more samples as tensor gets more biased.}
%\label{fig:plot56}
%\end{center}
%\vskip -0.2in
%\end{figure*}
%
%\begin{figure*}[ht]
%\vskip 0.2in
%\begin{center}
%\centerline{\includegraphics[width=120mm]{figures/noise}}
%\caption{ In this plot we consider the performance of algorithm~\ref{algo:approx} in the noisy tensor case and plot the RMSE error of the computed factors from algorithm~\ref{algo:approx} as the noise Frobenius norm increases. The underlying tensor is sum of random orthogonal tensor with entrywise Gaussian noise. As we can see the proposed sampling distribution has the small error rate as we increase the noise Frobenius norm.}
%\label{fig:plot7}
%\end{center}
%\vskip -0.2in
%\end{figure*}

In this section we present some simulation results comparing the proposed sampling technique to other distributions on synthetic examples. First we will present results for tensor sparsification followed by tensor completion and approximate factorization.\\

\noindent {\it Tensor sparsification:} We will now discuss the parameters of the simulations. We construct symmetric $100*100*100$ order 3-tensors. We generate $p$ random unit vectors $X_i$ and the corresponding tensor $\sum_{i=1}^p X_i \otimes X_i \otimes X_i$. We plot the error with the increasing number of samples $m$. Note that computing spectral norm of a tensor is NP-hard~\citep{hillar2013most}. Hence we use the following approximation of spectral norm as the error measure. $\|\T\|_{2,2}^2 = \sum_{i=1}^n \|\T_{i,:,:}\|^2$, which is 2-norm of spectral norm of each face of the tensor. Note that since the tensor is symmetric we can consider faces along any dimension.

We compare the error performance with the following distributions: uniform, L2: $\pijk \propto \T_{ijk}^2$, Sum L3: $\pijk \propto \|X^i\|^3 + \|X^j\|^3 + \|X^k\|^3$, and the proposed distribution Tensor L.S. $\pijk \propto ~\eqref{eq:tensor_ls}$.% Prod L3: $\pijk \propto ~\eqref{eq:samp_prod}$

In the figure~\ref{fig:plot12} we compare performance of various sampling distributions as we increase the number of samples. For this plot we create tensor from random samples, $\T=\sum_{i=1}^p X_i\otimes^3$ with $p=50$. For plot~\ref{fig:plot12}(a) we generate $X_i$ from random Gaussian vectors. For plot~\ref{fig:plot12}(b) we bias $X_i$ according to a power law with diagonal matrix $D_{ii} = \frac{1}{i^a}$, and use $D*X_i$. We set $a=0.5$. This generates tensors concentrated in fewer elements and hence uniform sampling trivially incurs more error. We see that the proposed sampling distribution has the smallest error as we increase the number of samples.

In figure~\ref{fig:plot34} we plot error performance as we increase number of sample vectors $p$ for various distributions. We fix the number of entries sampled from the tensor at $m =\lceil 10*n^{1.5} \rceil$. As we increase the number of vectors $p$ the approximation becomes worser and the error increases. Again in~\ref{fig:plot34}(a) we use tensor constructed from random vectors $X_i$ and most distributions have similar error. In ~\ref{fig:plot34}(b) we consider tensor constructed from biased random vectors $D*X_i$, with $a=0.5$ and we notice that the proposed sampling distribution has smaller error.\\

\noindent {\it Tensor completion:}
In figure~\ref{fig:plot56}(a) we plot the performance of algorithm~\ref{algo:wals}. We consider rank-5 orthogonal tensors $\T= \sum_{i=1}^5 U_i \otimes U_i \otimes U_i, U =SVD(D*X)$ with varying bias $a$, and plot the number of samples needed for exact recovery of various sampling techniques. We show that proposed sampling distribution (tensor L.S) needs (approximately) same number of samples irrespective of the bias ($a$) of the factors. Other distributions need increasingly more samples for recovery as the bias of the factors increases.\\

\noindent {\it Tensor factorization:} In figure~\ref{fig:plot56}(b) we plot the performance of algorithm~\ref{algo:approx}. We construct random orthogonal tensors with noise $\T= \sum_{i=1}^5 U_i \otimes U_i \otimes U_i +\E$, where $\E$ is an entrywise random Gaussian tensor. We compute RMSE of the recovered factors with the true factors and plot this on y-axis vs increasing norm of the noise $\|\E\|_F$, on the x-axis.  Again we notice that the proposed sampling distribution has smaller error compared to other distributions.\\

%\end{document}

%\section{Conclusion}
%In this paper we have discussed new ways to sample tensors for three different problems, tensor sparsification, completion and approximate factorization. With growing data sizes 

%{\it Open problems:} Tensor algorithms being relatively new field has many interesting open problems and possible future directions from here. There is interest in proving more settings where tensor can be factorized in polynomial time. Another interesting direction is to remove the dependence on the condition number in the proof of the weighted alternating least squares algorithm.

%%%%%%%%%%%%%%%%%%%%%%%%%%%%%%%%%%%%%%%%%%%%%%%%%%%%%%%%%%%%%%%%%%%%%%%%%%

\clearpage
%\newpage
%\bibliographystyle{abbrv}
\bibliographystyle{icml2015}
%\bibliography{matrix,tensor}

\clearpage
%\newpage
\appendix
\section{Concentration results}
In this section we will review the concentration results we will be using in our proofs.

\begin{lemma}[Bernstein's Inequality]\label{lem:bernstein}
Let $X_1,...X_n$ be independent scalar random variables. Let $|X_i| \leq L, \forall i ~w.p. ~1$. Then,
\begin{equation}
\prob{\lv\sum_{i=1}^n X_i -\sum_{i=1}^n \expec{X_i}\rv \geq t} \leq 2\exp\left(\frac{-t^2/2}{\sum_{i=1}^n \Var(X_i)+ Lt/3}\right). 
\end{equation}
\end{lemma}

\begin{lemma}[Matrix Bernstein's Inequality~\cite{tropp2012user}]\label{lem:mbernstein}
Let $X_1,...X_p$ be independent random matrices in $\rnn$. Assume each matrix has bounded deviation from its mean: \[ \|X_i - \expec{X_i}\| \leq L, \forall i ~w.p. ~1.\] Also let the variance be \[ \sigma^2 =\max \left\{ \lV \expec{\sum_{i=1}^p (X_i-\expec{X_i}) (X_i- \expec{X_i})^T \rV}, \lV \expec{\sum_{i=1}^p (X_i - \expec{X_i})^T (X_i - \expec{X_i})}\rV \right\}.\] Then,
\begin{equation}
\prob{\lV\sum_{i=1}^n \left(X_i - \expec{X_i}\right)\rV \geq t} \leq 2n\exp\left(\frac{-t^2/2}{\sigma^2 + Lt/3}\right). 
\end{equation}
\end{lemma}

%\noindent Recall the Shatten-$p$ norm of a matrix $X$ is \begin{equation*}\|X\|_p = \left(\sum_{i=1}^n \sigma_i(X)^p \right)^{1/p}. \end{equation*} $ \sigma_i(X)$ is the $i$th singular value of $X$. In particular for $p=2$, Shatten-$2$ norm is the Frobenius norm of the matrix.
%\begin{lemma}\label{lem:mchebyshev}[Matrix Chebyshev Inequality~\cite{mackey2012matrix}]
%Let $X$ be a random matrix. For all $t >0$,
%\begin{equation}
%\prob{\lV X \rV \geq t} \leq \inf_{p \geq 1} t^{-p} \expec{\lV X \rV_p^p}.
%\end{equation}
%\end{lemma}

Now from~\citep{nguyen2010tensor} we know the following bound on spectral norm of a random tensor.

%Let k-norm of a vector in $x \in \R^n$ be $\|x\|_k= \left(\sum_{i=1}^n \lv x_i \rv^k \right)^{\frac{1}{k}}$. We define the sampling distribution used. 
%\begin{equation}
%\pik =\frac{\sum_{l=1}^k \|X^{i_l}\|_k^k}{k \times n^{k-1} \times \sum_{i=1}^n \|X^{i}\|_k^k }.
%\end{equation}
%We sample $\ik$ entry of tensor $\T$ with probability $\phik =m *\pik$.

\begin{theorem}\label{thm:tensor_conc}
Let $\T \in  \R^{n\times \cdots \times n}$ be an order-$d$ tensor and let $\wT$ be a random tensor of same dimensions with independent entries such that $\expec{\wT} =\T$. For any $\lambda \leq \frac{1}{64}$, and $1\leq q \leq 2nd\lambda\ln (\frac{5e}{\lambda})$, then:
\begin{equation}\label{eq:tensor_conc}
\left(\expec{\|\wT -T\|^q_2}\right)^{\f1q} \leq ~ c 8^d \sqrt{2d \ln \left( \frac{5e}{\lambda}\right)} \left( \left[ \log_2 \left( \frac{1}{\lambda}\right) \right]^{d-1} \left(\sum_{j=1}^d \expec{\alpha_j^q} \right)^{\f1q} + \sqrt{\lambda n}  \expec{\beta^q}^{\f1q} \right),
\end{equation}
where 
\begin{equation*}
\alpha_j^2 \eqdef  \max_{i_1, \cdots i_{j-1}, i_{j+1} \cdots i_d} \left( \sum_{i_j =1}^ n \wT_{i_1 \cdots i_d}^2 \right) ~ \text{and} ~ \beta \eqdef \max_{i_1 \cdots i_d} \lv \wT_{i_1 \cdots i_d} \rv
\end{equation*}
\end{theorem}

%%%%%%%%%%%%%%%%%%%%%%%%%%%%%%%%%%%%%%%%%%%%%%%%%
%%%%%%%%%%%%%%%%%%%%%%%%%%%%%%%%%%%%%%%%%%%%%%%%%

\section{Proofs of Section~\ref{sec:tensor_spfcn}}\label{sec:apdx_init}

First we will give certain properties of the sampling distribution. Recall $$ p_{ijk}= \frac{\|X^i\|^3\|X^j\|^3 + \|X^j\|^3\|X^k\|^3 + \|X^k\|^3\|X^i\|^3}{3 n \|X\|_3^2},$$ where $\|X\|_3 = \sum_{i=1}^n \|X^i\|^3$. Also we will use $\dijk$ to denote indicator random variable throughout the proofs.

\begin{lemma}\label{lem:init_prob1} Given a tensor $\T=\sum_{i=1}^p X_i \otimes X_i \otimes X_i$ and distribution $\pijk$ defined in equation~\eqref{eq:samp_prod} the following holds,
\begin{equation}\label{eq:init_prob1} \frac{\T_{ijk}^2 }{\pijk} \leq  n \|X\|_3^2  ~\forall (i, j, k).\end{equation}
%\begin{equation}\label{eq:init_prob2}\frac{(\Uo_{jq})^2 (\Uo_{kq})^2 }{\pijk} \leq 3n \|\Uo\|_{3/2}^2 . \end{equation}
\end{lemma}

\begin{proof}
First using Cauchy-Schwarz inequality we get, $$\T_{ijk} =\sum_{l=1}^p X_{il} X_{jl} X_{kl} \leq \sqrt{ \sum_{l=1}^p  X_{il}^2} \sqrt{ \sum_{l=1}^p X_{jl}^2 X_{kl}^2} \leq  \|X^i\| \|X^j\|\|X^k\|.$$ Also by AM-GM inequality we get,  $$p_{ijk}=\frac{\|X^i\|^3\|X^j\|^3 + \|X^j\|^3\|X^k\|^3 + \|X^k\|^3\|X^i\|^3}{3 n \|X\|_3^2}  \geq \frac{\|X^i\|^2 \|X^j\|^2 \|X^k\|^2}{n \|X\|_3^2}.$$ Hence the first inequality follows from the above two equations.

%For proving second inequality we use the fact that $\|(\Uo)^j\| \leq 1$. Hence $$\frac{(\Uo_{jq})^2 (\Uo_{kq})^2 }{\pijk} \leq 3n \|\Uo\|_{3/2}^2 \frac{(\Uo_{jq})^2 (\Uo_{kq})^2 }{\|(\Uo)^j\|^{\ftt} \|(\Uo)^k\|^{\ftt}} \leq 3n \|\Uo\|_{3/2}^2$$.
\end{proof}

Now we will provide the proof of Theorem~\ref{thm:tensor_init}. We will use the relation between the tensor $\T$ and the probability distribution $\pijk$ through Lemma~\ref{lem:init_prob1}.

\begin{proof}[Proof of Theorem~\ref{thm:tensor_init}]

To bound $\lV \wT - \T \rV$ we will use the concentration theorem~\ref{thm:tensor_conc}. Let $\H = \wT - \T$. Now if $\widehat{p}_{ijk} \geq 1$, then $\H_{ijk}=\wT_{ijk} - \T_{ijk}=0$. Hence we only consider the cases for which $\widehat{p}_{ijk} =m*p_{ijk} \leq 1$.

We follow the same strategy of~\citep{nguyen2010tensor} to bound $\lV \wT - \T \rV$, by dividing the indices $(i, j, k)$ into various sets and bounding the error $\lV \wT - \T \rV$ over each set separately. %Let $\gamma = O(\frac{\eps^2}{n^{k/2}})$ and 
Let $\T^{[1]}$ denote tensor with only entries such that $\pijk \geq \frac{1}{2m}$ and similarly let $\T^{[l]}$ be the tensor with only entries of $\T$ satisfying $\pijk \in \left[ \frac{1}{2^l m}, \frac{1}{2^{l+1} m} \right)$. Similarly we define $\wT^{[l]}$ corresponding to sampled and rescaled entries of $\T^{[l]}$. Also let $s=\lceil \log (n^{3/2}/\ln^3 n) \rceil$. Hence using triangle inequality we get,
\[\lV \wT - \T \rV \leq \lV \wT^{[1]} - \T^{[1]} \rV + \sum_{l=2}^s \lV \wT^{[l]} - \T^{[l]} \rV +  \sum_{l=s+1} \lV \wT^{[l]} - \T^{[l]} \rV. \]

\noindent Now we will bound each of the above three terms in the summation.\\

\noindent {\bf $\mathbf{l=1}$ case:}

%Let $\H = \wT^{[1]} - \T^{[1]} $. Then $$ \H_{ijk}^2 \leq \frac{\T_{ijk}^2}{m^2 \pijk^2} \stackrel{\zeta_1}{\leq} \frac{n^2 \|X\|_3^4}{m^2 \T_{ijk}^2} \stackrel{\zeta_2}{\leq} \frac{2 n^2 \|X\|_3^4}{m}.$$ $\zeta_1$ follows from~\eqref{eq:init_prob1}. $\zeta_2$ follows from $\T_{ijk}^2 \geq \frac{1}{2m}$. Hence $\lv \H_{ijk} \rv \leq \sqrt{2}\frac{n \|X\|_3^2}{sqrt{m}}$.

Let $\H = \wT^{[1]} - \T^{[1]} $. Then $$ \H_{ijk}^2 \leq \frac{\T_{ijk}^2}{m^2 \pijk^2} \stackrel{\zeta_1}{\leq} \frac{n \|X\|_3^2}{m^2 \pijk} \stackrel{\zeta_2}{\leq} \frac{2 n \|X\|_3^2}{m},$$ where $\zeta_1$ follows from~\eqref{eq:init_prob1} and $\zeta_2$ follows from $\pijk \geq \frac{1}{2m}$. Hence $\lv \H_{ijk} \rv \leq \sqrt{\frac{2n \|X\|_3^2}{m}}$. This implies $$\max_{jk} (\sum_i \H_{ijk}^2)^{q/2} \leq n^{q/2} \left(\sqrt{\frac{2n \|X\|_3^2}{m}} \right)^q.$$ Now applying the tensor concentration theorem~\ref{thm:tensor_conc} for $\lambda =\frac{1}{64}$ and $q \leq 5n/8$ gives us, 
$$\left(\expec{\lV \H \rV^q}\right)^{1/q} \leq C 3^{1/q} \sqrt{n} \sqrt{\frac{n \|X\|_3^2}{m}}.$$\\

\noindent {\bf $\mathbf{l \leq \lceil \log (n^{3/2}/\ln^3 n) \rceil}$ case:}

Again let $\H = \wT^{[l]} - \T^{[l]} $. Then, $$ \H_{ijk}^2 \leq \frac{\T_{ijk}^2}{m^2\pijk^2} \leq  \frac{n \|X\|_3^2}{m^2 \pijk} \leq \frac{ 2^l n \|X\|_3^2}{m}.$$ Further $\expec{\max_{jk} \left(\sum_i \H_{ijk}^2 \right)^{q/2} } \leq \sqrt{\expec{\max_{jk} \left(\sum_i \H_{ijk}^2 \right)^q}}$.  Hence using the above two equations we get,

$$\expec{\max_{jk} \left(\sum_i \H_{ijk}^2 \right)^q} \leq  \left(\frac{2^l n \|X\|_3^2}{m}\right)^{q/2} \expec{\max_{jk} \left(\sum_i \dijk \right)^q}.$$

Now using Lemma 17 from~\citep{nguyen2010tensor} we know that $\expec{\max_{jk} \left(\sum_i \dijk \right)^q} \leq 2(5n2^{-l} + 6\ln (n) + 2q)^{q}$.  Hence from the above two equations and using theorem~\ref{thm:tensor_conc} gives us,
$$ \left( \expec{\lV \wT^{[l]} - \T^{[l]} \rV^q} \right)^{1/q} \\ \leq C \sqrt{6 \ln \left( \frac{5e}{\lambda}\right)} \left( \left[ \log_2 \left( \frac{1}{\lambda}\right) \right]^{2} 3^{1/q} \sqrt{5n +(3\ln(n)+q)*2^{l+1}}  + \sqrt{\lambda 2^l n} \right) \sqrt{\frac{n \|X\|_3^2}{m}}.$$

\noindent {\bf $\mathbf{l \geq \lceil \log (n^{3/2}/\ln^3 n) \rceil}$ case:}
For this case note that $\pijk \leq \frac{\ln^3 n}{m n^{3/2} }$. Hence from~\eqref{eq:init_prob1} $\T_{ijk}^2 \leq n \|X\|_3^2\frac{\ln^3 n}{m n^{3/2} } $. Since the elements of $\T$ are small in this case the error is also small. Hence,
$$\|\wT- \T\| \leq \sqrt{\sum_{ijk} \T_{ijk}^2} \leq \sqrt{n \|X\|_3^2 n^{3/2} \ln^3(n)/m}.$$

Applying Markov\rq{}s inequality with $q=6\ln(n)$, combining the above three bounds we get 
$$\lV \wT - \T \rV \leq C\frac{\sqrt{n^{3/2} \ln^3(n)}}{\sqrt{m} } \sqrt{n} \|X\|_3$$ with probability $\geq 1-\frac{1}{n^6}$.
\end{proof}

\section{Proofs of Section~\ref{sec:tensor_als}}\label{sec:app_2}
 In this section we will present the proof of theorem~\ref{thm:tensor_als}. First we will give certain properties of the sampling distribution. Recall $$ p_{ijk}= \frac{\|(\Uo)^i\|^{3/2} \|(\Uo)^j\|^{3/2}+ \|(\Uo)^j\|^{3/2} \|(\Uo)^k\|^{3/2} + \|(\Uo)^k\|^{3/2} \|(\Uo)^i\|^{3/2}}{3n \|\Uo\|_{3/2}^2},$$ where $\|\Uo\|_{3/2} = \sum_{i=1}^n \|(\Uo)^i\|^{3/2}$. Also we will use $\dijk$ to denote the indicator random variable throughout the proofs.

\begin{lemma}\label{lem:prob1}
\begin{equation}\label{eq:prob1} \frac{\T_{ijk}}{\pijk} \leq  n \so_{\max}\|\Uo\|_{3/2}^2 . \end{equation}
\begin{equation}\label{eq:prob2}\frac{(\Uo_{jq})^2 (\Uo_{kq})^2 }{\pijk} \leq 3n \|\Uo\|_{3/2}^2 . \end{equation}
\begin{equation}\label{eq:prob3}\frac{(\Uo_{iq}) (\Uo_{jq}) (\Uo_{kq}) }{\pijk} \leq n \|\Uo\|_{3/2}^2 . \end{equation}
\end{lemma}

\begin{proof}
Recall $$\T_{ijk} =\sum_{l=1}^r \so_l \Uo_{il} \Uo_{jl} \Uo_{kl} \leq \so_{\max} \sqrt{ \sum_{l=1}^r  (\Uo_{il})^2} \sqrt{ \sum_{l=1}^r(\Uo_{jl})^2 (\Uo_{kl})^2} \leq \so_{\max} \|(\Uo)^i\|\|(\Uo)^j\|\|(\Uo)^k\|.$$ Also by AM-GM inequality we get  $$p_{ijk}= \frac{\|(\Uo)^i\|^{3/2} \|(\Uo)^j\|^{3/2}+ \|(\Uo)^j\|^{3/2} \|(\Uo)^k\|^{3/2} + \|(\Uo)^k\|^{3/2} \|(\Uo)^i\|^{3/2}}{3n \|\Uo\|_{3/2}^2} \geq \frac{\|(\Uo)^i\| \|(\Uo)^j\| \|(\Uo)^k\|}{n \|\Uo\|_{3/2}^2}.$$ Hence the first inequality follows from the above two equations.\\

\noindent For proving second inequality we use the fact that $\|(\Uo)^j\| \leq 1$. Hence $$\frac{(\Uo_{jq})^2 (\Uo_{kq})^2 }{\pijk} \leq 3n \|\Uo\|_{3/2}^2 \frac{(\Uo_{jq})^2 (\Uo_{kq})^2 }{\|(\Uo)^j\|^{\ftt} \|(\Uo)^k\|^{\ftt}} \leq 3n \|\Uo\|_{3/2}^2.$$

\noindent The proof of third inequality follows from  $$p_{ijk}\geq \frac{\|(\Uo)^i\| \|(\Uo)^j\| \|(\Uo)^k\|}{n \|\Uo\|_{3/2}^2}.$$ 
\end{proof}

To show that the algorithm~\ref{algo:wals} recovers the underlying factors we show that each iteration decreases the distance to the true factors. For this we define the following notion of distance. Let $U_l$ and $\sigma_l$ be iterates at the end of iteration $t$. Then \begin{equation}\label{eq:dist} d_{\infty}([U, \Sigma], [\Uo, \So]) = \max_l (\|d_l\| + \Delta_l),\end{equation} where $\|d_l\| =\| U_l - \Uo_l\|$ and $\Delta_l =\frac{\lv \sigma_l - \so_l \rv}{\so_l}$.

Now we will show that the distance to iterates at the end of $t+1$ iteration decreases geometrically.

\begin{theorem}\label{thm:infty}
Let  $d_{\infty}([U, \Sigma], [\Uo, \So]) \leq \frac{1}{100*r*\kappa}$ and $U$ satisfies~\eqref{eq:init_coh}, then,
\begin{equation*}
d_{\infty}([U^{(t+1)}, \Sigma^{(t+1)}], [\Uo, \So]) \leq \frac{1}{2} d_{\infty}([U, \Sigma], [\Uo, \So]),
\end{equation*} 
with probability greater than $1-\frac{1}{n^{10}}$,~ for $m \geq O((\sum_i \|U^i\|^{\ftt})^2 nr^3\kappa^4  \log^2(n))$. Further $U^{(t+1)}$ satisfies~\eqref{eq:init_coh}.
\end{theorem}

\begin{proof}
Recall that $T=\sum_{l=1}^r \so_l \Uo_l \otimes \Uo_l \otimes \Uo_l$ and $\wutp_q= \argmin_{u \in \R^n}\| R_{\Omega_{t*r+q}}^{1/2}(\T- u \otimes U_q \otimes U_q -\sum_{l\neq q}\sigma_l U_l \otimes U_l \otimes U_l)\|_F^2$. Hence, \begin{align}\label{eq:wals_update} \wutp_{iq}  = \frac{\sum_{jk} \dijk \wijk \Uo_{jq} \Uo_{kq} U_{jq} U_{kq}}{\sum_{jk} \dijk \wijk U_{jq}^2 U_{kq}^2}\so_q \Uo_{iq} + \sum_{l \neq q}\frac{\sum_{jk} \dijk \wijk  (\so_l \Uo_{il} \Uo_{jl}\Uo_{kl} - \sigma_l U_{il} U_{jl} U_{kl}) U_{jq} U_{kq} }{\sum_{jk} \dijk \wijk U_{jq}^2 U_{kq}^2}.\end{align}

Now we will show that the distance between update $\wutp_q$ and $\Uo_q$ decreases with each iteration by expressing the update in terms of tensor power method update and error similar to the proof of~\citep{jain2014provable}. For the rest of the proof we will use the same notation as in~\citep{jain2014provable}. \begin{multline}\label{eq:wals_pupdate} \wutp_{q}  = \so_q \ip{\Uo_q}{U_q}^2 \Uo_q - B^{-1}(\so_q \ip{\Uo_q}{U_q}^2 B - \so_q C)\Uo_q \\+ \sum_{l \neq q} (\so_l \ip{\Uo_l}{U_l}^2\Uo_l - \sigma_l \ip{U_q}{U_l}^2 U_l) + \sum_{l \neq q} B^{-1} (\so_l (\ip{U_q}{\Uo_l}^2 B -F^{(l)})\Uo_l -\sigma_l (\ip{U_q}{U_l}^2 B -G^{(l)})U_l) ,\end{multline} where $B, C, F^{(l)}, G^{(l)}$ are diagonal matrices with,
\begin{align*}
B_{ii} = \sum_{jk} \dijk \wijk U_{jq}^2 U_{kq}^2,~~ &C_{ii} = \sum_{jk} \dijk \wijk U_{jq} \Uo_{jq} U_{kq} \Uo_{kq}, \\
F_{ii}^{(l)} = \sum_{jk} \dijk \wijk U_{jq} \Uo_{jl} U_{kq} \Uo_{kl},~ \text{and }~  &G_{ii}^{(l)} =\sum_{jk} \dijk \wijk U_{jq} U_{jl} U_{kq} U_{kl}.  
\end{align*}

Now define the error \begin{align}\label{eq:err_update}
&err_q^0 =  \so_q (\ip{\Uo_q}{U_q}^2-1) \Uo_q - B^{-1}(\so_q \ip{\Uo_q}{U_q}^2 B - \so_q C)\Uo_q \nonumber \\
&err_l^1  =   (\so_l \ip{\Uo_l}{U_l}^2\Uo_l - \sigma_l \ip{U_q}{U_l}^2 U_l) \nonumber \\
&err_l^2  =  B^{-1} (\so_l (\ip{U_q}{\Uo_l}^2 B -F^{(l)})\Uo_l -\sigma_l (\ip{U_q}{U_l}^2 B -G^{(l)})U_l)
\end{align}

The goal is to bound each of these error terms in terms of distances $\|d_l\|$ and $\Delta_l$ so as to express $d_{\infty}([U^{(t+1)}, \Sigma^{(t+1)}], [\Uo, \So])$ in terms of $d_{\infty}([U, \Sigma], [\Uo, \So])$ . Now we will bound the error  $\|\wutp_q - \so_q \Uo_q\| = err^0_q +\sum_{l \neq q}(err_l^1 + err_l^2)$. By Lemma~\ref{lem:supp3} and~\ref{lem:supp1} we get, $$err^0_q \leq \so_q (1-\ip{\Uo_q}{U_q}^2 +2\gamma \sqrt{1-\ip{\Uo_q}{U_q}^2}) \leq \so_q \|d_q\| (\|d_q\| +2 \gamma).$$

Using Lemma B.10 and B.11 of~\cite{jain2014provable} we get, $$\|err_l^1\| \leq 4\so_l (\|d_l\| +\|d_q\|) (\|d_l\| + \Delta_l),$$ and  
\begin{align*}
 \so_l (\ip{U_q}{\Uo_l}^2 B -G^{(l)})\Uo_l &-\sigma_l (\ip{U_q}{U_l}^2 B -F^{(l)})U_l =  (\ip{U_q}{\Uo_l}\ip{U_q}{d_l}B- D^{(1)})\Uo_l \\ &+  (\ip{U_q}{\Uo_l}\ip{U_q}{d_l}B- D^{(2)})\Uo_l +  (\ip{U_q}{d_l}^2 B- D^{(3)})\Uo_l-  (\ip{U_q}{U_l}^2 B- F^{(l)}) d_l \\
 &\stackrel{\zeta_1}{\leq} 8 \gamma \so_l \|d_l\|.
\end{align*}
Note that $D^{(i)}$ are diagonal matrices with $D^{(1)}_{ii} =\sum_{jk}\dijk \wijk U_{jq} U_{kq} d_l(j) \Uo_{kl}$,  $D^{(2)}_{ii} =\sum_{jk}\dijk \wijk U_{jq} U_{kq} d_l(k) \Uo_{jl}$ and  $D^{(3)}_{ii} =\sum_{jk}\dijk \wijk U_{jq} U_{kq} d_l(j) d_l(k)$. $\zeta_1$ follows from Lemma~\ref{lem:supp3}, Lemma~\ref{lem:supp4}. Hence $||err_l^2|| \leq 8\gamma \so_l (\|d_l\| + \Delta_l)$. Combining the bounds on all the error terms and setting $\gamma=\frac{1}{100r \kappa}$, we get $$\|\wutp_q- \so_q \Uo_q\| \leq \frac{\so_q}{16 \kappa} \|d_q\| + 1/16 \so_{\min}  d_{\infty}([U, \Sigma], [\Uo, \So]).$$\\

Further $|\sigma^{t+1}_q -\so_q| \leq |\sigma^{t+1}_q U^{t+1}_q -\so_q \Uo_q| \leq \frac{\so_q}{8} d_{\infty}([U, \Sigma], [\Uo, \So])$ and $\so_q\| \utp_q -\Uo_q\| \leq \frac{\so_q}{4} d_{\infty}([U, \Sigma], [\Uo, \So])$. Hence combining these two equations we get, $$\|d_q^{t+1}\| + \Delta_q^{t+1}  \leq  \frac{1}{2}d_{\infty}([U, \Sigma], [\Uo, \So]).$$

\noindent Now we will prove the second part of the theorem.
\begin{align*}
\lv \wutp_{iq}\rv &\leq \so_q \frac{\lv C_{ii} \rv}{\lv B_{ii} \rv}\Uo_{iq} + \sum_{l \neq q} \so_l \frac{\lv F_{ii}^{(l)} \rv}{\lv B_{ii} \rv}\Uo_{il} +\sum_{l \neq q} \so_l \frac{\lv G_{ii}^{(l)} \rv}{\lv B_{ii} \rv}\Uo_{il} \\
&\leq \|(\Uo)^i\| \left( \so_q \frac{1+\gamma}{1-\gamma} +  \sum_{l \neq q} \so_l (\gamma +\|d_l\|) + \sum_{l \neq q} \so_l (1 + \Delta_l) (\gamma + \|\delta_l\|) \right)\\
&\leq\|(\Uo)^i\| \so_q (1+1/100),
\end{align*}
since $\gamma \leq \frac{1}{100r\kappa}$. Using the bound on $|\sigma^{t+1}_q -\so_q|$ from above the result follows.
\end{proof}

\begin{proof}[Proof of Theorem~\ref{thm:tensor_als}]
The proof now follows from Theorem~\ref{thm:infty}. After $\log(4\sqrt{r}\|T\|_F/\eps)$ iterations, the error $\|U_q - \Uo_q\| \leq \frac{\eps}{4\sqrt{r}\|\T\|_F}$ and $\lv \sigma_q -\so_q \rv \leq \frac{\eps}{4\sqrt{r}\|\T\|_F}$. Hence from Lemma 2.4 of~\citep{jain2014provable} it follows that $\|\wT - \T\| \leq \|\wT - \T\| _F \leq \eps$.
%\begin{align*}
%&\|\wT - \T\| \leq \|\wT - \T\| _F \\
%&\|
%\end{align*}
\end{proof}

%%%%%%%%%%%%%%%%%%%%%%%%%%%%%
\subsection{Supporting lemmas}

\begin{lemma}\label{lem:supp1}
For $\Omega$ generated according to~\eqref{eq:tensor_ls} and $U$ satisfying~\eqref{eq:init_coh}, there exists a constant $C$ such that the following holds:
\begin{equation*}  \lv \sum_{jk} \dijk \wijk U_{jq} \Uo_{jq} U_{kq} \Uo_{kq} - \ip{U_q}{\Uo_q}^2\rv \leq \gamma, \end{equation*} 
for any fixed $q$, with probability greater than $1-\frac{1}{n^{10}}$, for $m \geq \frac{C}{\gamma^2} n \log(n)  \|\Uo\|_{3/2}^2.$
\end{lemma}

\begin{proof}
Let $X_{jk} =\dijk \wijk U_{jq} \Uo_{jq} U_{kq} \Uo_{kq}$. From~\eqref{eq:init_coh} we get $|X_{jk}| \leq \wijk 4 \frac{\|(\Uo)^j\|^2 \|(\Uo)^k\|^2}{\|\Uo\|_F^2} \leq \frac{4 3n \|\Uo\|_{3/2}^2}{m }$. Also,
$$\expec{\sum_{jk} X_{jk}^2 } \leq \sum_{jk}\wijk (U_{jq} \Uo_{jq} U_{kq} \Uo_{kq} )^2  \leq \frac{ 3n \|\Uo\|_{3/2}^2}{m}\sum_{jk} U_{jq}^2 U_{kq}^2 =\frac{ 3n \|\Uo\|_{3/2}^2}{m}.$$

Hence by applying the Bernstein's inequality the result follows.
\end{proof}

\begin{lemma}\label{lem:supp2}
For $\Omega$ generated according to~\eqref{eq:tensor_ls} and $U$ satisfying~\eqref{eq:init_coh}, there exists a constant $C$ such that the following holds for any fixed $b \in \R^n$:
\begin{equation*} \lv \sum_{jk} \dijk \wijk \Uo_{iq} U_{jq} \Uo_{jq} U_{kq} b_k- \Uo_{iq} \ip{U_q}{\Uo_q}\ip{U_q}{b} \rv \leq \gamma \|b\|, \end{equation*} 
with probability greater than $1-\frac{1}{n^{10}}$, for $m \geq \frac{C}{\gamma^2} n \log(n) \|\Uo\|_{3/2}^2.$
\end{lemma}

\begin{proof}
Let $X_{jk} =\dijk \wijk \Uo_{iq} U_{jq} \Uo_{jq} U_{kq} b_k$. From~\eqref{eq:prob3} and \eqref{eq:init_coh} we get $|X_{jk}| \leq  \frac{12 n \|U\|_{3/2}^2 \|b\|}{m}$. Also,
$$\expec{\sum_{jk} X_{jk}^2 } \leq \sum_{jk}\wijk (\Uo_{iq} U_{jq} \Uo_{jq} U_{kq} \Uo_{kq} b_k )^2  \leq \frac{12 n \|U\|_{3/2}^2}{m}\sum_{jk} U_{jq}^2 b_k^2 =\frac{ 12n \|U\|_{3/2}^2 \|b\|^2}{m}.$$

Hence by applying the Bernstein's inequality the result follows.
\end{proof}

\begin{lemma}\label{lem:supp3}
Let $\Omega$ be generated according to~\eqref{eq:tensor_ls}, $U$ satisfying~\eqref{eq:init_coh} and fixed unit vectors $a, b$ and $c$ in $\R^n$ such that $|a_i|, |b_i|, |c_i| \leq 2\|(\Uo)^i\| \forall i$. Let $B$ and $R$ be diagonal matrices with $B_{ii}=\sum_{jk} \dijk \wijk U_{jq}^2 U_{kq}^2$ and $R_{ii}= \sum_{jk} \dijk \wijk U_{jq} U_{kq} a_j b_k$. Then, there exists a constant $C$ such that the following holds:
\begin{equation*}  \|(\ip{U_q}{a}\ip{U_q}{b}B - R)c \| \leq \gamma\sqrt{1-\ip{U_q}{a}^2\ip{U_q}{b}^2}, \end{equation*} 
for any fixed $q$, with probability greater than $1-\frac{1}{n^{10}}$, for $m \geq \frac{C}{\gamma^2} n \log(n) \|\Uo\|_{3/2}^2$.
\end{lemma}

\begin{proof}
Let $X_{ijk} = \dijk \wijk c_i U_{jq} U_{kq} (U_{jq} U_{kq} \ip{U_q}{a}\ip{U_q}{b} - a_j b_k ) e_i$. Note that $\sum_{jk}\expec{X_{ijk}}=0$.  $$\|X_{ijk}\| \stackrel{\zeta_1}{\leq} \frac{8 n* \|\Uo\|_{3/2}^2}{m} \sqrt{\sum_{jk}(U_{jq} U_{kq} \ip{U_q}{a}\ip{U_q}{b} - a_j b_k )^2} = \frac{8n* \|U\|_{3/2}^2}{m}  \sqrt{1- \ip{U_q}{a}^2 \ip{U_q}{b}^2}.$$ $\zeta_1$ follows from~\eqref{eq:prob3}.

Also, $$\|\sum_{ijk}\expec{X_{ijk}^T X_{ijk}}\| = \lv \sum_{ijk} \wijk c_i^2 U_{jq}^2 U_{kq}^2 (U_{jq} U_{kq} \ip{U_q}{a}\ip{U_q}{b} - a_j b_k )^2\rv \stackrel{\zeta_1}{\leq} \frac{ 48 n \|\Uo\|_{3/2}^2}{m} (1- \ip{U_q}{a}^2 \ip{U_q}{b}^2).$$ $\zeta_1$ follows from~\eqref{eq:prob2}. Hence the lemma follows from applying the matrix Bernstein inequality.
\end{proof}

\begin{lemma}\label{lem:supp4}
Let $\Omega$ be generated according to~\eqref{eq:tensor_ls}, $U$ satisfying~\eqref{eq:init_coh} and fixed unit vectors $a, b$ and $c$ in $\R^n$ such that $|a_i|, |b_i|, |c_i| \leq 2\|(\Uo)^i\|, \forall i$. Let $B$ and $R$ be diagonal matrices with $B_{ii}=\sum_{jk} \dijk \wijk U_{jq}^2 U_{kq}^2$ and $R_{ii}= \sum_{jk} \dijk \wijk U_{jq} U_{kq} a_j b_k$. Then, there exists a constant $C$ such that the following holds:
\begin{equation*}  \|(\ip{U_q}{a}\ip{U_q}{b}B - R)c \| \leq \gamma \|b\|, \end{equation*} 
for any fixed $q$, with probability greater than $1-\frac{2}{n^{9}}$, for $m \geq \frac{C}{\gamma^2} n \log(n) \|\Uo\|_{3/2}^2$.
\end{lemma}

\begin{proof}
Let $X_{ijk} =\dijk \wijk c_i U_{jq} U_{kq} a_j b_k e_i $. Then $\sum_{jk}\expec{X_{ijk}} = c_i \ip{U_q}{a}\ip{U_q}{b}$. $$\lv X_{ijk} \rv \leq  \frac{12 n \|U\|_{3/2}^2 \|b\|}{m}.$$ Further $$\|\sum_{ijk}\expec{X_{ijk}^T X_{ijk}}\| = \lv \sum_{ijk} \wijk c_i^2 U_{jq}^2 U_{kq}^2 a_j^2 b_k^2 \rv \leq \frac{12 n \|U\|_{3/2}^2 \|b\|^2}{m}.$$ Hence $\|(\ip{U_q}{a}\ip{U_q}{b} - R)c \| \leq \gamma \|b\|$ from matrix Bernstein\rq{}s inequality. 

From Lemma~\ref{lem:supp1} $|B_{ii}| \leq 1+\gamma$. Hence applying union bound over all $i$ we get $\|B-I\| \leq \gamma$. Hence the lemma follows.

% and from Lemma~\ref{lem:supp2} $\ip{U_q}{a}\ip{U_q}{b}B_{ii}c_i - R_{ii}c_i \leq 2\gamma \|b\|$ with probability $\geq 1-\frac{1}{n^{10}}$. Hence the lemma follows by union bound over all $i$, since  $\|(\ip{U_q}{a}\ip{U_q}{b}B - R)c \| = \max_{i}  |\ip{U_q}{a}\ip{U_q}{b}B_{ii}c_i - R_{ii}c_i|$.

\end{proof}

%%%%%%%%%%%%%%%%%%%%%%%%%%
\subsection{Initialization}

From Theorem 5.1~\citep{anandkumar2014tensor} we know that Robust Tensor Power Method (RTPM) gives a good approximation of factors for small error.

\begin{lemma}\label{lem:app_init}
Let $\|\Ro(\T) -\T \| \leq \delta$, then $c \log(r)$ iterations of RTPM on $\Ro(\T)$ achieves:\begin{align*}
\|U_l -\Uo_l\| \leq c \kappa r \delta, ~\text{and} \lv \sigma_l -\so_l\rv \leq \so_l \kappa r \delta, \end{align*}
with probability greater than $1-1/n^5$, for all $l \in [r]$.
\end{lemma}

We further threshold entries of $U$ such that $U_{il} \leq 2 \|(\Uo)^i\|$. Note that we can estimate these quantities from the samples. Hence this guarantees that initialization satisfies \begin{equation}\label{eq:init_coh}  \lv U_{il}\rv \leq 2 \|(\Uo)^i\|. \end{equation}

%\begin{lemma}\label{lem:estim_thresh}
%For
%\end{lemma}

%%%%%%%%%%%%%%%%%%%%%%%%%%%%%%%
%%%%%%%%%%%%%%%%%%%%%%%%%%%%%%%
\section{Proofs of Section~\ref{sec:approx}}

In this section we will present the proof of Theorem~\ref{thm:approx}. The proof follows the same way as the noiseless version with key modifications which we will discuss now. We will first present the key properties of the sampling distribution which we will use in the rest of the proof.
 Recall $ \pijk = 0.5\frac{\nu_i^{\ftt} \nu_j^{\ftt} +\nu_j^{\ftt} \nu_k^{\ftt} + \nu_k^{\ftt} \nu_i^{\ftt}}{3nZ} +0.5 \frac{\T_{ijk}^2}{\|\T\|_F^2},$ where $\nu_i = \frac{\|T_{i, :, :}\|_F}{\|\T\|_F} + \frac{1}{\sqrt{n}}$ and $Z= \left(\sum_{i=1}^n \nu_i^{\ftt}\right)^2 $ is the normalizing constant. Recall $\T_{ijk} = \sum_{l=1}^r \so_l \Uo_{il} \Uo_{jl} \Uo_{kl} + \E_{ijk}$.

\subsection{Initialization}
First we will show that sampling~\eqref{eq:samp_approx} followed by RTPM generates a good approximation to the underlying factors. 
\begin{lemma}\label{lem:approx_init}
Given $\T = \sum_{l=1}^r \so_l \Uo_{l} \otimes \Uo_{l}\otimes \Uo_{l} + \E$ where $\Uo$ is orthonormal matrix and $\E$ satisfies~\eqref{eq:noise_assume}, the output $U$ of step 4 of algorithm~\ref{algo:approx} satisfies the following:
\begin{equation*}
\|U_i -\Uo_i\| \leq \frac{1}{100r \kappa} ~~\text{and }~~ |U_{ij}| \leq 2\nu_i,  \forall i \in [r],
\end{equation*}
with probability $\geq 1-\frac{1}{n^5}$ for $m \geq O(n^{1.5} r^3 \log^3(n) \kappa^4)$.
%let $\Ro(\T)$ be the sampled and reweighed tensor according to distribution~\eqref{eq:samp_approx}
\end{lemma}

\begin{proof}
First from Theorem 1 of~\citep{nguyen2010tensor} we get that $\|\Ro(\T) -\T\| \leq \eps\|\T\|_F$, for $m \geq O(\frac{n^{1.5}}{\eps^2} \log^3(n))$. Note that by triangle inequality, and equation~\eqref{eq:noise_assume} we get, $$\|\T\|_F \leq \sqrt{\sum_{i=1}^r (\so_i)^2} + \|\E\|_F \leq 2 \so_{\max} \sqrt{r}.$$ Hence, $$\|\Ro(\T) -\sum_{l=1}^r \so_l \Uo_{l} \otimes \Uo_{l} \otimes \Uo_{l}\| \leq 2 \eps \so_{\max} \sqrt{r} + \|\E\| \leq  \eps(2 \so_{\max} \sqrt{r}) + C\frac{\so_{\min}}{100 r} \leq C\frac{\so_{\min}}{r},$$ for $\eps \leq \frac{C}{r^{1.5} \kappa}$, which is true for $m \geq O(n^{1.5} r^3 \kappa^4 \log^3(n))$. Hence the factors $U$ computed using RTPM on $\Ro(\T)$ satisfies $$ \|U_l - \Uo_l \| \leq   \frac{1}{100\kappa r}, ~ \forall l \in [r],$$ from Theorem 5.1 of~\citep{anandkumar2014tensor}, for $m \geq  O(n^{1.5} r^3\kappa^4 \log^3(n) )$. %we can guarantee initialization satisfies $$  \|U_i -\Uo_i\| \leq \frac{1}{100r\kappa}.$$ 

Further we threshold each entry of $U$ such that $U$ satisfies~\eqref{eq:init_approx}. Note that the proof that thresholding step doesn\rq{}t increase the distance to the optimal factors by more than a constant factor, follows the same way as in proof of Lemma 3.2 in~\citep{doi:10.1137/1.9781611973730.62} and we will not discuss it here.\\
\end{proof}

\subsection{WALS}
Now before we present the proof of Theorem~\ref{thm:approx}, we will present some bounds on the error because of the noise in each stage of the WALS algorithm. 

%Now we will present the proof for the second part of the algorithm~\ref{algo:approx}. 
One key modification compared to the noiseless case is, we need iterates to satisfy the following bound in each iteration. \begin{equation}\label{eq:init_approx}  \lv U_{ij}\rv \leq 2 \left( \frac{\|T_{i, :, :}\|_F}{\|\T\|_F} + \frac{1}{\sqrt{n} }\right), ~ \forall (i, j). \end{equation}\\

Now we will discuss some key properties of the sampling.

\begin{lemma}\label{lem:approx_prob1}
%\begin{equation}\label{eq:approx_prob1} \frac{\T_{ijk} }{\pijk} \leq  n \so_{\max}\sqrt{r}\|U\|_{3/2}^2 . \end{equation}
For $U$ satisfying~\eqref{eq:init_approx} the following holds for distribution~\eqref{eq:samp_approx}.
\begin{equation}\label{eq:approx_prob2}\frac{(U_{jq})^2 (U_{kq})^2 }{\pijk} \leq 96 n Z . \end{equation}
\begin{equation}\label{eq:approx_prob3}\frac{(U_{iq}) (U_{jq}) (U_{kq}) }{\pijk} \leq 16n Z . \end{equation}
\end{lemma}$Z$ is the normalizing constant in~\eqref{eq:samp_approx}.

\begin{proof}
%Recall  \begin{multline*} \T_{ijk} =\sum_{l=1}^r \so_l \Uo_{il} \Uo_{jl} \Uo_{kl} + \E_{ijk} \leq \so_{\max} \sqrt{ \sum_{l=1}^r  (\Uo_{il})^2} \sqrt{ \sum_{l=1}^r (\Uo_{jl})^2 (\Uo_{kl})^2} + \so_{\max}/n \\ \leq \so_{\max} \|(\Uo)^i\| \|(\Uo)^j\|\|(\Uo)^k\|+\so_{\max}/n .\end{multline*} Also by AM-GM inequality we get  $$p_{ijk}= 0.5\frac{\nu_i^{\ftt} \nu_j^{\ftt} +\nu_j^{\ftt} \nu_k^{\ftt} + \nu_k^{\ftt} \nu_i^{\ftt}}{3nZ} +0.5 \frac{\T_{ijk}^2}{\|\T\|_F^2} \geq 0.5 \frac{\nu_i\nu_j\nu_k}{3n Z}.$$ Hence the first inequality follows from the above two equations.

The proof follows the same way as proof of Lemma~\ref{lem:prob1}.
\end{proof} 

Since most of the supporting lemmas in Section~\ref{sec:app_2} depend on the relations above, they all follow immediately. Next we will characterize the error by noise in each iteration of WALS.
\begin{lemma}\label{lem:noise_bound}
For $\E$ satisfying~\eqref{eq:noise_assume} and iterate $U$ satisfying~\eqref{eq:init_approx} the following holds,
\begin{equation*} \|\sum_i \sum_{jk}\dijk \wijk \E_{ijk}U_{jq} U_{kq} e_i -\sum_i \sum_{jk} \wijk \E_{ijk}U_{jq} U_{kq} e_i \| \leq \eps \|\E\|_F.\end{equation*}
with probability $\geq 1-\frac{1}{n^{10}}$, for $m \geq O( \frac{n Z}{\eps^2} \log^2(n))$. 
\end{lemma}

\begin{proof}
Let $X_{ijk} = \dijk \wijk \E_{ijk}U_{jq} U_{kq} e_i$. Then,
\begin{align*} \lV X_{ijk} \rV \stackrel{\zeta_1}{\leq}  \frac{ C\sqrt{ nZ} \E_{ijk}}{m \sqrt{\pijk}} \stackrel{\zeta_2}{\leq} \frac{ C\sqrt{ nZ} \|\E\|_F}{m n \sqrt{\pijk}} \stackrel{\zeta_3}{\leq} \frac{C nZ \|\E\|_F}{m}. \end{align*}
$\zeta_1$ follows from~\eqref{eq:approx_prob2}. $\zeta_2$ follows from~\eqref{eq:noise_assume}. $\zeta_3$ follows from $\pijk \geq \frac{1/n^{3/2}}{2nZ}$. Now we will bound the variance. 
\begin{align*}
\lV \sum_{ijk}\expec{X_{ijk}^T X_{ijk}} \rV &= \sum_i \sum_{jk} \wijk \E_{ijk}^2U_{jq}^2 U_{kq} ^2\\
& \stackrel{\zeta_1}{\leq} \sum_i \sum_{jk} \frac{\E_{ijk}^2 * CnZ }{m} = \frac{\|\E\|_F^2* CnZ}{m}.
\end{align*}
$\zeta_1$ follows from~\eqref{eq:approx_prob2}. Hence by matrix Bernstein\rq{}s inequality, the Lemma follows.
\end{proof}

Note that $$\lV \sum_i\sum_{jk} \wijk \E_{ijk}U_{jq} U_{kq} e_i \rV = \lV \sum_i U_q^T (\E_{i,:,:} )U_q e_i \rV \leq \|\E\|.$$ Hence the above lemma implies $\|\sum_i \sum_{jk}\dijk \wijk \E_{ijk}U_{jq} U_{kq} e_i\| \leq \|\E\| + \eps \|\E\|_F$, w.h.p.  \\

\begin{lemma}\label{lem:approx_infty}
Let  $d_{\infty}([U, \Sigma], [\Uo, \So]) \leq \frac{1}{100*r*\kappa}$, $U$ satisfies~\eqref{eq:init_approx} and $\E$ satisfies~\eqref{eq:noise_assume} then,
\begin{equation*}
d_{\infty}([U^{(t+1)}, \Sigma^{(t+1)}], [\Uo, \So]) \leq \frac{1}{2} d_{\infty}([U, \Sigma], [\Uo, \So]) +\frac{6\|\E\| + \eps \|\E\|_F}{\so_{\min}},
\end{equation*} 
with probability greater than $1-\frac{1}{n^{10}}$,~ for $m \geq O( \frac{nZ}{\eps^2} r^3\kappa^4  \log^2(n))$. Further $U^{(t+1)}$ satisfies~\eqref{eq:init_approx}.
\end{lemma}

\begin{proof}
The proof follows the same line as proof of Theorem~\ref{thm:infty}. Hence we only discuss the modifications caused from the noiseless case by the additional noise term.

 \begin{multline}\label{eq:wals_approx} \wutp_{iq}  = \frac{\sum_{jk} \dijk \wijk \Uo_{jq} \Uo_{kq} U_{jq} U_{kq}}{\sum_{jk} \dijk \wijk U_{jq}^2 U_{kq}^2}\so_q \Uo_{iq} + \sum_{l \neq q}\frac{\sum_{jk} \dijk \wijk  (\so_l \Uo_{il} \Uo_{jl}\Uo_{kl} - \sigma_l U_{il} U_{jl} U_{kl}) U_{jq} U_{kq} }{\sum_{jk} \dijk \wijk U_{jq}^2 U_{kq}^2} \\ + \frac{\sum_{jk}\dijk \wijk \E_{ijk}U_{jq} U_{kq} }{\sum_{jk} \dijk \wijk U_{jq}^2 U_{kq}^2}.\end{multline}

From Lemma~\ref{lem:noise_bound} and~\ref{lem:supp1} we get the following bound on the norm of noise term in each iteration, $$\|\sum_i \frac{\sum_{jk}\dijk \wijk \E_{ijk}U_{jq} U_{kq} }{\sum_{jk} \dijk \wijk U_{jq}^2 U_{kq}^2} e_i \| \leq  2(\|\E\| + \eps \|\E\|_F).$$ Hence $$\|\wutp_q- \so_q \Uo_q\| \leq \frac{\so_q}{16 \kappa} \|d_q\| + 1/16 \so_{\min}  d_{\infty}([U, \Sigma], [\Uo, \So]) + 2(\|\E\| + \eps \|\E\|_F).$$

Further $|\sigma^{t+1}_q -\so_q| \leq |\sigma^{t+1}_q U^{t+1}_q -\so_q \Uo_q| \leq \frac{\so_q}{8} d_{\infty}([U, \Sigma], [\Uo, \So]) +2(\|\E\| + \eps \|\E\|_F)$ and $\so_q\| \utp_q -\Uo_q\| \leq \frac{\so_q}{4} d_{\infty}([U, \Sigma], [\Uo, \So])+4(\|\E\| + \eps \|\E\|_F)$. Hence combining these two equations we get, $$\|d_q^{t+1}\| + \Delta_q^{t+1}  \leq  \frac{1}{2}d_{\infty}([U, \Sigma], [\Uo, \So])+\frac{6\|\E\| + \eps \|\E\|_F}{\so_q}.$$

\noindent Now we will prove the second part of the theorem.
\begin{align*}
\lv \wutp_{iq}\rv &\leq \so_q \frac{\lv C_{ii} \rv}{\lv B_{ii} \rv}\Uo_{iq} + \sum_{l \neq q} \so_l \frac{\lv F_{ii}^{(l)} \rv}{\lv B_{ii} \rv}\Uo_{il} +\sum_{l \neq q} \so_l \frac{\lv G_{ii}^{(l)} \rv}{\lv B_{ii} \rv}\Uo_{il} \\
&\leq \|(\Uo)^i\| \left( \so_q \frac{1+\gamma}{1-\gamma} +  \sum_{l \neq q} \so_l (\gamma +\|d_l\|) + \sum_{l \neq q} \so_l (1 + \Delta_l) (\gamma + \|\delta_l\|) \right)\\
&\leq\|(\Uo)^i\| \so_q (1+1/100),
\end{align*}
since $\gamma \leq \frac{1}{100r\kappa}$. Using the bound on $|\sigma^{t+1}_q -\so_q|$ from above the result follows.

%  $\eps \|\E\|_F$ to the error in Theorem~\ref{thm:infty}. By conditions on $\E$~\eqref{eq:noise_assume} we get that the distance decreases in each iteration.

Further to show that the iterates satisfy conditions~\eqref{eq:init_approx} consider the following.
$$\lv \wutp_{iq}\rv \leq \|(\Uo)^i\| \so_q(1+1/100) + \sum_{jk} \dijk \wijk \E_{ijk} U_{jl} U_{kl} \leq 
2\so_{\max} (\frac{\|T_{i, :, :}\|_F}{\|\T\|_F}+ \frac{1}{\sqrt{n}}).$$ 

To bound $\sum_{jk} \dijk \wijk \E_{ijk} U_{jl} U_{kl}$, note that $$\lv \wijk \E_{ijk} U_{jl} U_{kl} \rv \leq \frac{C nZ\so_{\min}}{r n^{1.5} m \sqrt{\pijk}} \leq \frac{C nZ}{m} \frac{\so_{\min}}{rn^{0.75}}$$ and $$\sum_{jk} \wijk \E_{ijk}^2 U_{jl}^2 U_{kl}^2 \leq \frac{CnZ}{m} \sum_{jk} \E_{ijk}^2 \leq \frac{CnZ}{m} \frac{ (\so_{\min})^2}{r n}.$$ Hence with high probability by Bernstein\rq{}s inequality we can say that $$\sum_{jk} \dijk \wijk \E_{ijk} U_{jl} U_{kl} \leq \sum_{jk} \E_{ijk} U_{jl} U_{kl} +\frac{\so_{\min}}{\sqrt{n}} \leq (1+\frac{1}{100})\frac{\so_{\min}}{\sqrt{n}}.$$\\

Hence the result follows.
\end{proof}

Note that for $m \geq O(\frac{n^{1.5}}{\eps^2} r^3\kappa^4 \log^3(n) )$, the error in the above lemma decreases from $\eps \frac{\|\E\|_F}{\so_{\min}}$ to $\eps\frac{\|\E\|_F}{\so_{\min}} \frac{\sqrt{Z}}{n^{0.25}}$. Now we have all the ingredients to present the proof of Theorem~\ref{thm:approx}.\\

\begin{proof}[Proof of Theorem~\ref{thm:approx}]
The proof now follows from Lemma~\ref{lem:approx_infty}. After $\log(4\sqrt{r}\|T\|_F/\gamma)$ iterations,  $\|U_q - \Uo_q\| \leq \frac{\gamma}{4\sqrt{r}\|\T\|_F} +\|\E\| \frac{12}{\so_{\min}} +\eps\|\E\|_F \frac{1}{\so_{\min}}$ and $\lv \sigma_q -\so_q \rv \leq \frac{\gamma}{4\sqrt{r}\|\T\|_F} +\|\E\|\frac{12\so_q}{\so_{\min}} +\eps \|\E\|_F\frac{\so_q }{\so_{\min}} $. 

Hence, $$\|\sum_l \sigma_l U_l \otimes^3 - \sum_l \so_l \Uo_l \otimes^3\| \leq  \sum_l \lv\sigma_l - \so_l \rv + \sum_l \so_l \| U_l \otimes^3 -  \Uo_l \otimes^3\|.$$ 
\begin{align*}
\| U_l \otimes^3 -  \so_l \Uo_l \otimes^3\| &\leq \| (U_l -\Uo_l) \otimes U_l \otimes U_l\| + \|U_l \otimes (U_l -\Uo_l) \otimes \Uo_l\|+ \| \Uo_l \otimes \Uo_l \otimes (U_l -\Uo_l)\| \\
& \leq 3 \|U_l -\Uo_l \|.
\end{align*}
Hence combining the above two relations we get, \begin{align*}\|\sum_l \sigma_l U_l \otimes^3 - \sum_l \so_l \Uo_l \otimes^3\| &\leq 4 \sum_l \so_l \left(\frac{\gamma}{4\sqrt{r}\|\T\|_F} +\|\E\| \frac{12}{\so_{\min}} +\eps\frac{\|\E\|_F}{\so_{\min}} \frac{\sqrt{Z}}{n^{0.25}}\right) \\
&\leq \gamma + 48r\kappa \|\E\| + \eps \|\E\|_F\frac{\sqrt{Z}}{n^{0.25}}.
\end{align*}
\end{proof}

%Now to bound $\sum_{jk} \dijk \wijk \E_{ijk} U_{jl} U_{kl}$, notice that 
%$$\lv\wijk \E_{ijk} U_{jl} U_{kl}\rv \leq \lv \frac{\E_{ijk}}{\sqrt{\pijk}} \leq 

% $\|\wutp_q- \so_q \Uo_q\| \leq \frac{\so_q}{16 \kappa} \|d_q\| + 1/16 \so_{\min}  d_{\infty}([U, \Sigma], [\Uo, \So]) +$.

\end{document}